\documentclass[journal]{IEEEtai}

\usepackage[colorlinks,urlcolor=blue,linkcolor=blue,citecolor=blue]{hyperref}

\usepackage{color,array}
\usepackage{adjustbox}

\usepackage{pdflscape} 
\usepackage{graphicx}
\usepackage{float}
\usepackage[linewidth=1pt]{mdframed}
\usepackage{afterpage}
\usepackage{longtable}
\usepackage{url}
\usepackage{caption}
\usepackage{epstopdf}

\usepackage{xcolor}
\usepackage{rotating}

\usepackage[colorinlistoftodos]{todonotes}
\presetkeys{todonotes}{color=blue!20}{}

\usepackage{multirow}

\usepackage{array}
\newcolumntype{C}{ >{\centering\arraybackslash} m{1.0cm} }
\newcolumntype{Q}{ >{\centering\arraybackslash} m{3 cmcm} }
\newcolumntype{M}[1]{>{\centering\arraybackslash}m{#1}}
\newcolumntype{P}[1]{>{\centering\arraybackslash}p{#1}}
\begin{document}

\title{A Comprehensive Review of Transformer-based language models for Protein Sequence Analysis and Design}

\author{Nimisha Ghosh,
       Daniele Santoni,
       Debaleena Nawn,
       Eleonora Ottaviani
       and~Giovanni Felici
\thanks{N. Ghosh is with the Department
of Computer Science and Engineering, Shiv Nadar University, Chennai, Tamil Nadu India e-mail: (nimishaghosh@snuchennai.edu.in).}
\thanks{D. Santoni, E. Ottaviani and G. Felici are with Institute for System Analysis and Computer Science “Antonio Ruberti”, National Research Council of Italy, Rome, Italy}
\thanks{D. Nawn is with the Department of Computer Science and Engineering, Adamas University , West Bengal}
\thanks{Manuscript received April 19, 2005; revised August 26, 2015.}}
\markboth{Journal of \LaTeX\ Class Files,~Vol.~14, No.~8, August~2015}%
{Ghosh \MakeLowercase{\textit{et al.}}: Bare Demo of IEEEtran.cls for IEEE Journals}

\maketitle
\begin{abstract}
The impact of Transformer-based language models has been unprecedented in Natural Language Processing (NLP). The success of such models has also led to their adoption in other fields including bioinformatics. 
Taking this into account, this paper discusses recent advances in Transformer-based models for protein sequence analysis and design. 
In this review, we have discussed and analysed a significant number of works pertaining to such applications. These applications encompass gene ontology, functional and structural protein identification, generation of \textit{de novo} proteins and binding of proteins. We attempt to shed light on the strength and weaknesses of the discussed works to provide a comprehensive insight to readers. Finally, we highlight shortcomings in existing research and explore potential avenues for future developments.
We believe that this review will help researchers working in this field to have an overall idea of the state of the art in this field, and to orient their future studies.
\end{abstract}
\begin{IEEEImpStatement}
Understanding the full potential of Transformer-based models to use them in a fruitful way in bioinformatics is the need of the hour. This review provides a thorough and critical examination of Transformer-based language models in the context of protein sequence analysis and design, unifying recent advancements in this domain. By synthesizing current methodologies and identifying key limitations, the paper offers a roadmap for leveraging Transformer architectures and accelerate the use of such protein language models.  This survey further outlines pressing research gaps and proposes future directions, positioning the work as a foundational resource for interdisciplinary innovation in protein science.
\end{IEEEImpStatement}
\begin{IEEEkeywords}
Bioinformatics, Protein Design, Protein Sequences, Natural Language Processing, Transformers
\end{IEEEkeywords}

\maketitle		
\section{Introduction}
Starting from the 1940s, scientists began to study the amino acid composition of proteins in order to characterize tissues and species by their amino acid frequencies \cite{BEACH1943}. Insulin was the first protein to be fully sequenced in 1951 and 1952 (chains of bovine insulin B and A, respectively) by Sanger \cite{Sanger49} that led to him winning the Nobel prize in chemistry (1958). In the following years, as new technologies were developed, many other
sequences became available thereby opening new frontiers in Biology and Chemistry. In fact, in the 70s a substantial number of sequences was made available, and several new issues related to protein sequences arose. Besides the simple analysis of amino acid frequencies \cite{SMITH1966}, scientists tried to infer functional and structural features of proteins from amino acid sequence patterns, their contexts and their combinatorics properties. Many works showed that classification methods were capable of separating proteins into different families \cite{Ferran1991, Orengo1993,blekas2005, Exarchos2006,Kocsor2005}) based exclusively on their sequences. 

In recent times, the influence of Natural Language Processing (NLP) in the field of bioinformatics is unprecedented. Neural networks have been applied to NLP tasks for over two decades, with word embeddings like Word2Vec \cite{Mikolov2013a} playing a key role in text representation. Word2Vec, using either Continuous Skip-Gram (CSG) or Continuous Bag-of-Words (CBOW), learns word vectors by predicting context from a target word or vice versa.
The development of deep learning further advanced language modelling through architectures such as RNNs, Bi-RNNs \cite{Schuster1997}, LSTMs \cite{Hochreiter1997}, and GRUs \cite{cho2014}, which encode sequences into fixed-length vectors. However, these models often struggled with the vanishing gradient problem, which hindered effective training on long sequences.
To address this, Transformer models \cite{NIPS2017} were introduced, replacing recurrent structures with a multi-head attention mechanism (see Section 2). Unlike RNNs, Transformers process sequences in parallel and capture word dependencies without relying on past hidden states, effectively mitigating gradient issues and improving performance on long-range dependencies. 

The applications of machine learning and deep learning, especially Transformers, in bioinformatics have brought about a huge change. A protein sequence can be viewed as a collection of overlapping/non-overlapping subsequences of a given length, that can be treated as words in a sentence, drawing a parallel between natural language processing (NLP) and protein sequences. Just as in human language, where the words are formed from letters, proteins can be represented as sequences of words constructed from the 20 amino acids.
This kind of representation allows to apply NLP and Transformer-based architectures to protein sequences. 
In this regard, many Transformer-based models have been applied to nucleotide sequences \cite{ghosh2025}. In the nucleotide context, genome classification, starting from New Generation Sequencing (NGS) reads to species typing are often the major focus, while in the context of proteins the focus is on three-dimensional structure prediction or protein-protein binding as well as generation of synthetic proteins.    
 Some significant milestones achieved in the context of protein research for Transformer-based models are outlined in Table~\ref{tab1}.
\begin{table}\scriptsize
\centering
\begin{tabular}{|l|l|p{2.5cm}|}
\hline
\textbf{Year} & \textbf{Model}  & \textbf{Focus} \\
\hline
2019 & TAPE~\cite{rao2019} & Protein sequence classification and prediction \\
\hline
2020-2021 & AlphaFold \cite{Senior2020}, AlphaFold2~\cite{jumper2021} & Protein structure prediction  \\
\hline
2022 & ProtBERT~\cite{Elnaggar2022} & Protein sequence function prediction \\
\hline
2022 & ProteinBERT~\cite{Brandes2022} & Protein function and structure prediction \\
\hline
2021-2022 & ESM-models~\cite{Rives2021, Meier2021, Zlin2023} & Protein structure and function prediction \\
\hline
2020-2023 & ProGen \cite{Madani2020} and ProGen2~\cite{nijkamp2023} & Novel and viable protein sequence generation\\
\hline
2022 & ProtGPT2~\cite{Ferruz2022}, ESM-3~\cite{Hayes2025} & \textit{De novo} protein sequence generation\\\hline

\end{tabular}
\caption{Timeline of Transformer-based Models for Protein Sequences}
\label{tab1}
\end{table}
It is to be noted that almost all modern Transformer-based protein models use 1-mer (single amino acid) tokenization to preserve biological granularity. 
In the recent past, few reviews have been published based on protein structure prediction and protein design considering Transformers. 
In this regard, \cite{HUANG2023} has put forth algorithmic modelling techniques for protein structure prediction especially based on deep neural networks, where the authors have summarised works that cover MSA Transformer \cite{Rao2021}, RoseTTAFold \cite{Baek2021}, ProteinBERT \cite{Brandes2022}, RGN2 \cite{Chowdhury2022} etc.; all based on some form of Transformers. Dhanuka et al. \cite{Dhanuka2023}
have also put forth a survey based on the applications of deep learning for the prediction of protein functions. In this regard, they have provided an insight into the interpretability of the models used in such prediction. Le \cite{Le2023} has also discussed the recent progress of Transformer-based NLP models in proteome bioinformatics to showcase their advantages, limitations and possible applications for improving their accuracy and efficiency. The review also provides an idea of the various challenges and the future directions of using the discussed models. Use of large scale Transformer models for predicting protein characteristics has been explored in \cite{Chandra2023}.  The review in~\cite{LEE2025} examines the applications of large language models, particularly the ones based on Transformers, across protein design. The protein language models (pLMs) discussed in this work include ESM-based models, ProtGPT2, ProGen and ProGen2, PepMLM etc. 
Wang et al.~\cite{wang2025} have also provided a comprehensive review of the pLMs encompassing the model architectures, positional encoding, scaling laws as well as the datasets. However, Wang et al. provides a macro perspective of the works being done in pLMs.


The work presented here particularly addresses the use of Transformer-based models for protein sequence analysis and design, focusing on their strengths and weaknesses, providing the readers with a clear classification of works according to their biological significance and results.
Although the aforementioned works have done a commendable job in addressing the works in protein science, some of them focus narrowly on specific applications of Transformers within protein research — for example, protein function prediction~\cite{Dhanuka2023} or protein design~\cite{LEE2025}. Others, such as Wang et al.~\cite{wang2025} adopt a broader view, including large language models beyond Transformers, or they consider data types beyond protein sequences.
The review by Le \cite{Le2023} adopts an approach similar to ours by organizing studies according to biological categories. However, our review extends this categorization to additional areas such as Gene Ontology and Binding. Moreover, given the rapid growth of literature in this field, with a large number of Transformer-related papers published in the past two years, our review provides a significant updated and novel perspective.

It is relevant to note here that the effectiveness of Transformers on protein sequence analysis is largely affected by the quality of the data sets that are studied and used for training. One issue in this regard is the annotation of sequences and protein functions, that are often inferred by computational techniques and not from direct assays. Noises in the data may lead the model to encode incorrect relationships and drive to wrong conclusions. At the same time, data may contain biases: annotated data sets are often skewed towards the well-studied organisms, and the trained models may limit the quality of their results on novel proteins.
The flexibility of Transformer-based models, when properly exploited and tuned, provides the best tools to contrast data quality problems, in particular the presence of noise in the data; as for biases, the problem may be mitigated by experimental design and careful interpretation of the results. However, noise still remains as one of the main risks in the application of Machine Learning and Artificial Intelligence in this field. 
In this review we have limited the analysis to studies where the quality of the data used can be retained good according to current standards, and where the methods used to assess the validity and the generalization capability of the models obey the good practice in the field.
Before discussing the different contributions from the literature, we provide a brief overview of Transformers so that the readers are well equipped with such knowledge before diving into the application areas. We also reiterate here that we have considered only those works which are based on Transformers and not other deep learning models, and that the focus of this work is amino acid sequences and not nucleotide sequences.


\section{Transformers}
\begin{figure*}
	\centerline{
		\includegraphics[height=4.0in,width=6.0in]{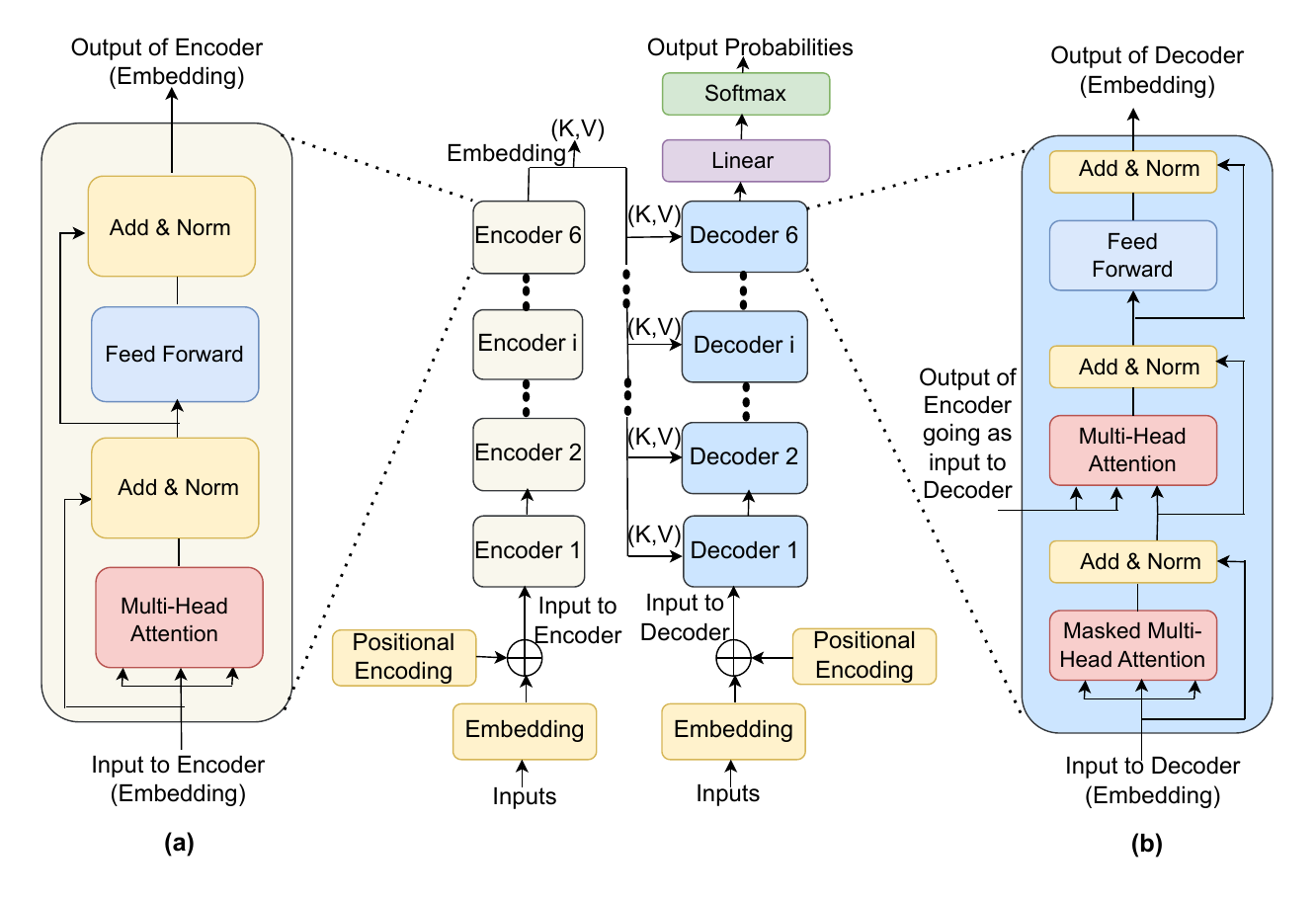}}
	\caption{Basic architecture and components of a Transformer: (a) Simplified scheme of an Encoder, (b) Simplified scheme of a Decoder, (c) Transformer architecture made of 6 such Encoders and Decoders stacked together wherein the outputs of the 6th Encoder (K and V matrices) are provided as inputs to each of the Decoder.
Considering an example of basic question answering, the input is a question which goes through different embeddings and then is provided as an input to Encoder 1 and sequentially to the other Encoders. The Decoder input is the embedding of the answer wherein the final output of the 6th Decoder passes through a linear layer and then through a softmax function to produce the output probabilities of the next predicted token. It is to be noted that each encoder and decoder has the same architecture and the zooming on Encoder 6 and Decoder 6 is just for representation and understanding purpose.
}
	\label{fig1}
\end{figure*}

Transformers have revolutionized Natural Language Processing (NLP) and are now being widely adapted in bioinformatics due to the sequential nature of biological data, such as DNA, RNA, and protein sequences. The main innovation introduced by Transformers is the self-attention mechanism. At a high level, a Transformer consists of two main components: the \textit{encoder} and the \textit{decoder}, each built from stacked layers containing two core sub-components; multi-head self-attention, which helps the model focus on relevant parts of the sequence, and position-wise feed-forward networks, applied independently to each token. Both of these sub-components are followed by residual connections and layer normalization, which aid in stable and efficient training. As shown in Figure~\ref{fig1}, multiple encoder and decoder layers are stacked together. Each encoder processes the entire input sequence and passes its output to the decoder, which generates the output sequence token-by-token in an autoregressive manner. The following subsections provide a concise idea about Transformers. For detailed descriptions, the readers may refer to~\cite{NIPS2017} and~\cite{ghosh2025}.

\subsection{Input Embedding and Positional Encoding}
Transformers process input tokens in parallel, unlike recurrent networks. Since this removes the natural order of sequences, positional encodings are added to word or token embeddings to capture token positions. The sum of the embedding and positional encoding forms the input to the first encoder layer. For biological sequences, similar tokenization can be employed.

\subsection{Attention Mechanism}
\begin{figure}
	\centerline{
		\includegraphics[height=3.2in,width=3.8in]{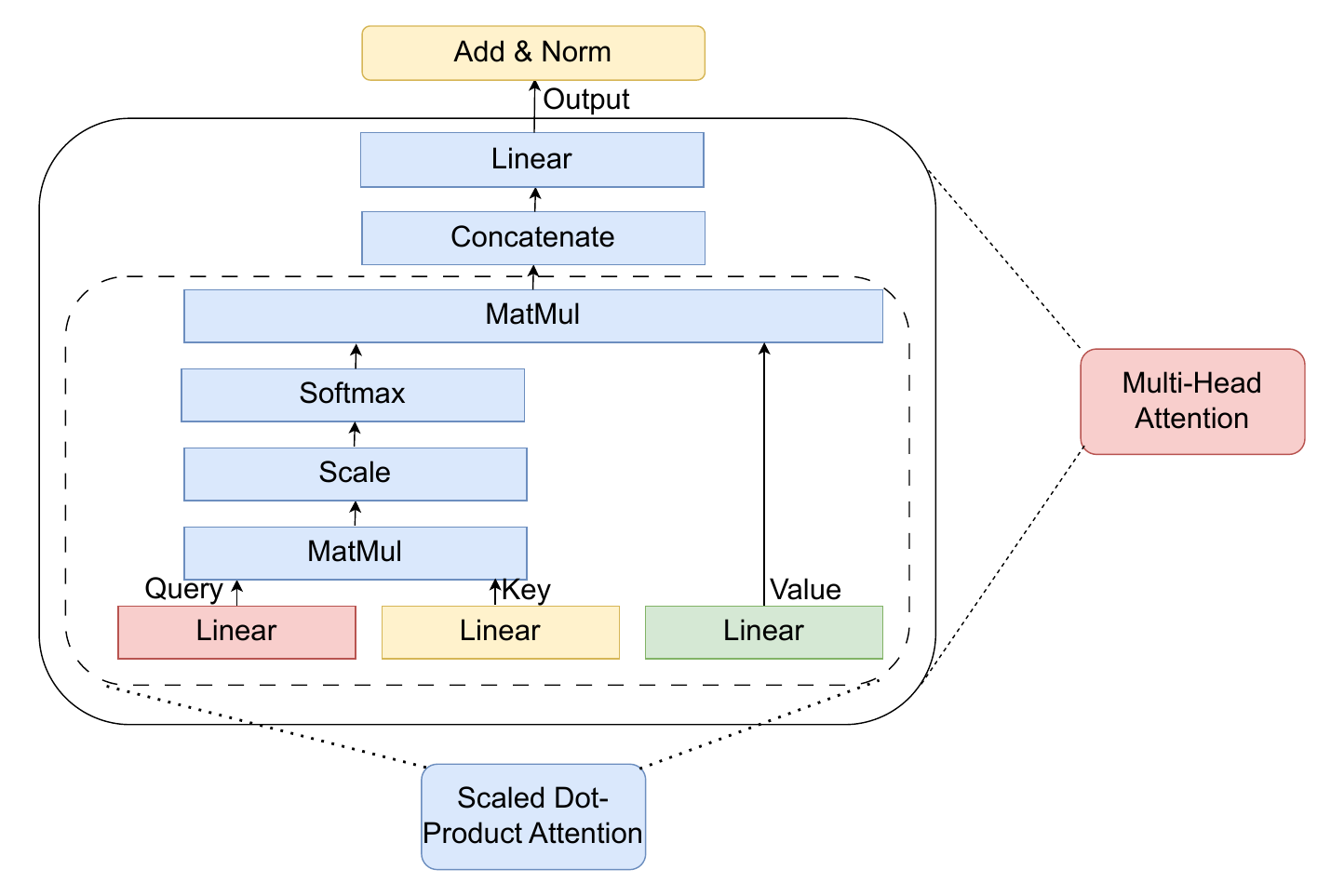}}
	\caption{Self-Attention in Transformer}
	\label{fig2}
\end{figure}

The core innovation in Transformers is the self-attention mechanism, where each token learns to focus on other relevant tokens in the sequence. The several components is shown in Figure \ref{fig2}.

Each input token is transformed into a query ($Q$), key ($K$), and value ($V$) vector through learned linear layers. Attention is computed using:
\begin{equation}
    Attention(Q, K, V) = softmax(\frac{QK^T}{\sqrt{d_k}})V
\end{equation}
This lets the model weigh the importance of each token in relation to others. In multi-head attention, multiple such attention heads are computed in parallel, and their outputs are concatenated and linearly transformed to capture diverse dependencies:
\begin{equation}
    MultiHead(Q, K, V) = Concat(head_1,\dots,head_h)W^O
\end{equation}
Here, $head_i$ = $Attention(QW_i^Q, KW_i^K, VW_i^V)$; $W^O$ is the random weight matrix that allows to perform a linear transformation (Linear in Figure \ref{fig2}). The concatenated matrix that results from the multiple head attentions is subject to this linear transformation, providing the output for multi-head attention layer of a Transformer.

\subsection{Feed-forward Networks and Normalization}
Apart from attention sub-layers, each of the encoder and decoder blocks contains a fully connected feed-forward network (depicted in Panels (a) and (b) of Figure~\ref{fig1}). This layer contains two linear transformations (two linear layers) with rectified linear unit (ReLU) activation in between.  

\begin{equation}
    FFN(x) = max(0,xW_1+b_1)W_2 + b_2
\end{equation}

Here, $W_1$, $b_1$, $W_2$ and $b_2$ are all learnable parameters.
Each of the encoder also contains two residual connection and normalisation layers. They are applied on both multi-head self attention and feed forward network, and calculated as follows:
 
\begin{equation}
    LayerNorm(X+MultiHead(X))
\end{equation}
\begin{equation}
    LayerNorm(X+FFN(X))
\end{equation}

where, $X$ is the input of multi-head self-attention/feed forward network, and is added to their respective outputs to form a residual connection.

\subsection{Encoder and Decoder}
The encoder processes the entire input sequence using stacked attention and feed-forward layers while the decoder generates output tokens one at a time, using both the encoder’s output and previously generated tokens. To prevent the decoder from accessing future tokens, a masking mechanism is applied within the self-attention layer to ensure that predictions are based only on past outputs. This results in autoregressive generation, where each new token is conditioned on the sequence generated so far. The decoder output is passed through a linear layer followed by a softmax to produce a probability distribution over possible next tokens, iteratively generating the output sequence until an end-of-sequence token is predicted.
This modular design, particularly the attention mechanism, makes Transformers highly versatile, enabling them to model complex dependencies in both language and biological sequences. Their parallelizable structure and ability to capture global context make them well-suited for tasks like sequence classification, mutation prediction, and functional annotation in bioinformatics. 
It is important to mention here that while the original Transformer architecture consists of both an encoder and a decoder, BERT~\cite{devlin2019} is based solely on the encoder stack of the Transformer. BERT uses a bidirectional approach, allowing each token to attend to both its left and right context simultaneously. It is pretrained using a masked language modelling (MLM) objective, where some input tokens are randomly masked and the model is trained to predict them using surrounding context. In contrast, GPT~\cite{radford2018} is based exclusively on the decoder stack of the Transformer and uses a unidirectional (left-to-right) causal language modelling objective, predicting the next token in a sequence given its left context. Both BERT and GPT have significantly influenced protein sequence modelling. Many methods discussed in the next section are built on BERT, GPT, or their respective variants, leveraging their ability to produce rich, contextual embeddings for diverse downstream tasks in protein sequence analysis. While BERT-based models are typically used for discriminative tasks such as gene ontology prediction or binding site identification, GPT-based models are increasingly applied to generative tasks such as \textit{de novo} protein design. For example, ESM-2~\cite{Zlin2023} is one of the more popular BERT-based models which is trained on hundreds of millions of protein sequences which captures deep evolutionary and structural patterns, producing high-quality embeddings useful for tasks like binding site prediction. ProtGPT2~\cite{Ferruz2022}, on the other hand, is a GPT-based autoregressive language model designed for \textit{de novo} protein sequence generation. Trained on large-scale protein sequence datasets, it generates syntactically valid and biologically plausible proteins by predicting amino acids left to right. ProtGPT2 emphasizes novelty and diversity, making it useful for protein design tasks.

\section{Applications of Transformer-based language models for Protein sequences}
In this section, a detailed analysis of different applications of Transformer-based language models for protein sequences is provided along with their main advantages and disadvantages. 
The considered studies are partitioned broadly into four areas, according to the biological context they refer to and analysed accordingly: Gene Ontology, Functional and Structural Protein Cluster Identification, Generating \textit{de novo} Proteins and Protein Binding. It is to be noted that some studies can fall into more than one category, as biological concepts such as function and structure are closely interconnected; for instance, Gene Ontology is inherently related to function, and the design of new functional proteins is inevitably tied to their function.
However, in most cases, the considered studies can be primarily associated with a single category. For example, if a study focuses on designing new functional proteins, it can be classified under \textit{de novo} proteins, even if it also involves proteins assigned to a particular functional class.
In instances where a work is strongly relevant to more than one category, we have identified  the most appropriate or natural classification. Moreover, in addition to their biological context, the models can also be categorised according to their architecture, a summary of which is provided in Table~\ref{model}.

\begin{table*}
\centering
\begin{tabular}{l|p{2cm}|p{3cm}|p{2cm}|p{5cm}}
\hline
\textbf{Architecture Type} & Category & \textbf{Model}  & {Generative or Discriminative} & {Pretrained or Finetuned} \\
\hline
 & Gene Ontology & OntoProtein~\cite{zhang2022ontoprotein} &  Discriminative & Pretrained (based on ProBert~\cite{Elnaggar2022}), further finetuned for downstream tasks\\\cline{3-5}
             && GO Models \cite{Vu2022} & Discriminative & Pretrained embeddings (ProtBert) and Finetuned (ProteinBERT~~\cite{Brandes2022})\\\cline{3-5}
           & & Zhao et al. \cite{Zhao2023bio} & Discriminative & Pretrained embeddings (ESM-1b~\cite{Rives2021})\\\cline{3-5}
          &&  GALA~\cite{Yfu2024} & Discriminative & Pretrained embeddings (ESM-1b)\\\cline{2-5}
         & Functional and Structural  & TooT-BERT-M \cite{Ghazikhani2022} & Discriminative & Finetuned (ProtBERT-BFD~\cite{Elnaggar2022})\\\cline{3-5}
          &  Protein Cluster Identification & TooT-BERT-C \cite{Ghazikhani2023} & Discriminative & Finetuned (ProtBERT-BFD and TooT-BERT-M)\\\cline{3-5}
           &  &  LMPhosSite \cite{Pakhrin2023} & Discriminative & Pretrained embeddings (ProtT5~\cite{Elnaggar2022}) \\\cline{3-5}
        && CaLMPhosKAN~\cite{Pratyush2025} & Discriminative & Pretrained embeddings (Codon adaptation Language Model and ProtT5)\\\cline{3-5}
        && DeepZF \cite{Gazit2022} & Discriminative & Finetuned (ProteinBERT)\\\cline{3-5} 
       & & LMNglyPred \cite{Pakhrin2023_1} & Discriminative & Pretrained embeddings (ProtT5)\\\cline{3-5}
         && DeepLoc-2.0 \cite{Thumuluri2022} &  Discriminative & Pretrained embeddings (ESM-2~\cite{Zlin2023}, ESM-1b, ProtT5)\\\cline{3-5}
        & & Adaptor \cite{Rahardja2022} & Discriminative & Neither (Uses Transformer encoder block as part of the proposed model) \\\cline{3-5}
        & & DAttProt \cite{Lin2022_1}& Discriminative & Pretrained (based on Transformer Encoder and  BERT-styled Masked LM), further finetuned for downstream tasks \\\cline{3-5}
      Encoder-Only & & PD-BertEDL \cite{Wang2022_1} & Discriminative & Pretrained embeddings (BERT-mini~\cite{turc2019})\\\cline{3-5}

     &  & MTL \cite{An2022} & Discriminative & Finetuned (BERT~\cite{devlin2019})\\\cline{3-5} 
         && SPRoBERTa \cite{Wu2022} & Discriminative & Pretrained (based on RoBERTa~\cite{liu2019}), further finetuned for downstream tasks\\\cline{3-5} 
         & & DistilProtBert \cite{Geffen2022}&  Discriminative & Pretrained (based on ProtBert), further finetuned for downstream tasks\\\cline{3-5}
          && Erckert et al.~\cite{Erckert2024} & Discriminative & Pretrained embeddings (SeqVec \cite{Heinzinger2019}, ProtBert, ProtT5)\\\cline{3-5}
        &  & TopLapGBT~\cite{WEE2024}& Discriminative & Pretrained embeddings (MSA Transformer)\\\cline{3-5} 
        & &PTSP-BERT~\cite{LV2025}&Discriminative & Pretrained embeddings (BERT-bfd~\cite{Elnaggar2022})\\\cline{3-5} 
        &&VUS model~\cite{Joshi2025} & Discriminative & Pretrained embeddings (ESM-2)\\\cline{2-5} 
       
         & Generating \textit{de novo} proteins & PeTriBERT \cite{Dumortier2022}& Generative & Trained from scratch (based on BERT)\\\cline{2-5}
    & Binding & DeepHomo2.0 \cite{Lin2022}&  Discriminative & Pretrained embeddings (MSA Transformer)\\\cline{3-5}
   &&  DTI-BERT \cite{Zheng2022} & Discriminative & Pretrained embeddings (ProtBert)\\\cline{3-5}
    & & TUnA~\cite{Ko2024} & Discriminative & Pretrained embeddings (ESM-2) \\\cline{3-5}
    & & LBCE-XGB \cite{Liu2023} & Discriminative & Pretrained embeddings (\cite{YZhang2021})\\\cline{3-5}
    && IDBindT5 \cite{Jahn2024} & Discriminative & Pretrained embeddings (ProtT5)\\\hline  
 &     Generating \textit{de novo} proteins & ProGen~\cite{Madani2020} & Generative & Trained from scratch (based on Transformer architecture), further finetuned for generating novel and functional protein sequences \\\cline{3-5}
   Decoder-Only  & &ProGen2 \cite{nijkamp2023}& Generative & Pretrained (based on Transformer Decoder), further finetuned for specific tasks\\\cline{3-5}
     & & ProtGPT2 \cite{Ferruz2022} & Generative & Pretrained (Based on Transformer Decoder), further finetuned for specific protein families \\\hline
     & Functional and Structural Protein Cluster Identification & MFTrans~\cite{CHEN2024} & Discriminative & Pretrained embeddings (MSA Transformer)~\cite{Rao2021} and Transformer architecture \\\cline{3-5} 
     &&ProsT5~\cite{Heinzinger2024} & Both & Pretrained (based on ProtT5), further finetuned for translating between protein sequences and structures  \\\cline{2-5}
 & Generating \textit{de novo} proteins  & xTrimoPGLM \cite{Chen2023}& Both & Pretrained (based on General Language Model (GLM)~\cite{Du2022}), further finetuned for downstream tasks\\\cline{3-5}
& &CFP-GEN~\cite{yin2025} & Generative & Trained from scratch (based on~\cite{XWang2024}), further finetuned for downstream tasks \\\cline{3-5}
Encoder-decoder& & Regression Transformer \cite{Born2023}  & Both & Pretrained (based on Transformer architecture), further finetuned for downstream tasks\\\cline{2-5}

& Binding &Trans-MoRFs~\cite{Meng2025} & Discriminative & Trained from scratch (based on Transformer architecture), further finetuned for downstream task\\\cline{3-5}

  &  & AppendFormer and MergeFormer~\cite{Wu2024} & Generative & Trained from scratch (based on Transformer architecture) \\\cline{3-5}
   & &GeoDock~\cite{ChuRu2024} &  Generative & Pretrained embeddings (ESM-2)\\\hline
\end{tabular}
\caption{Summary of the models based on their architecture}
\label{model}
\end{table*}

\subsection{Gene Ontology}
\begin{figure*}
				
				\centerline{
					\includegraphics[height=2.5in,width=3.0in]{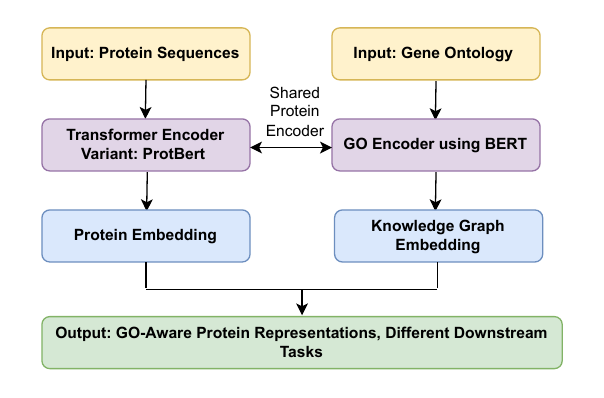}\hspace{2cm}
					\includegraphics[height=3.0in,width=2.7in]{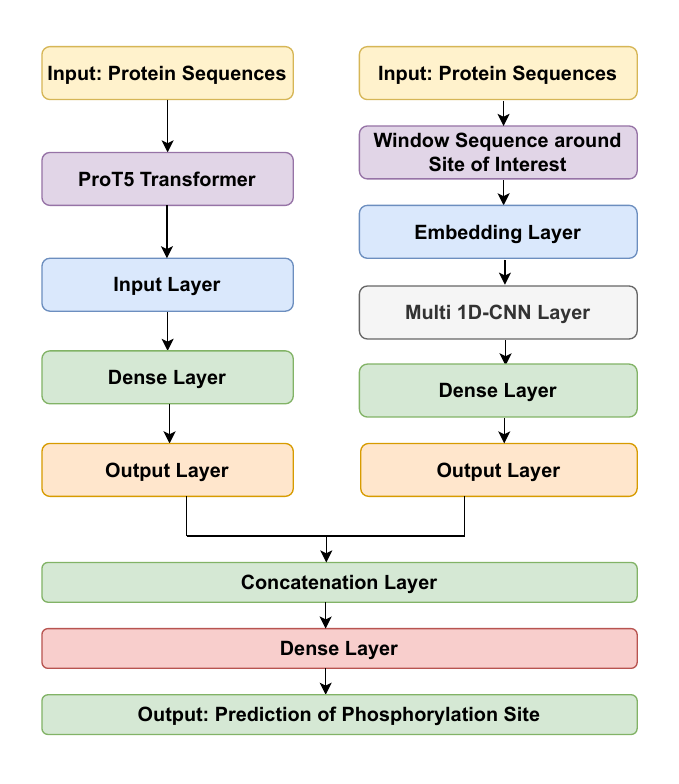}
				}\centerline{(a)\hspace{95mm}(b)}
			        \centerline{
				\includegraphics[height=3.0in,width=3.0in]{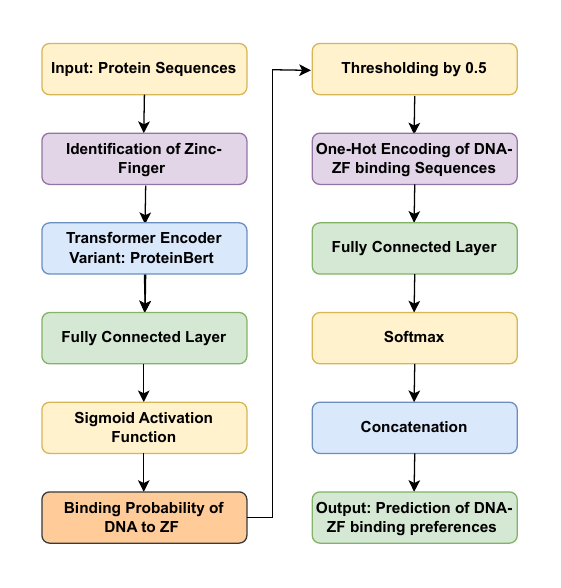}\hspace{2cm}
				 \includegraphics[height=3.0in,width=2.7in]{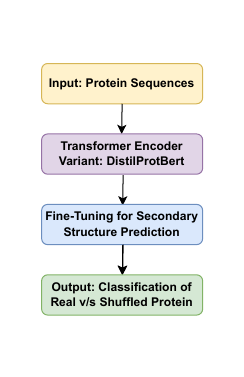}\hspace{0.2mm}}
				\centerline{(c)\hspace{95mm}(d)}
				\centerline{
                   \includegraphics[height=3.0in,width=3in]{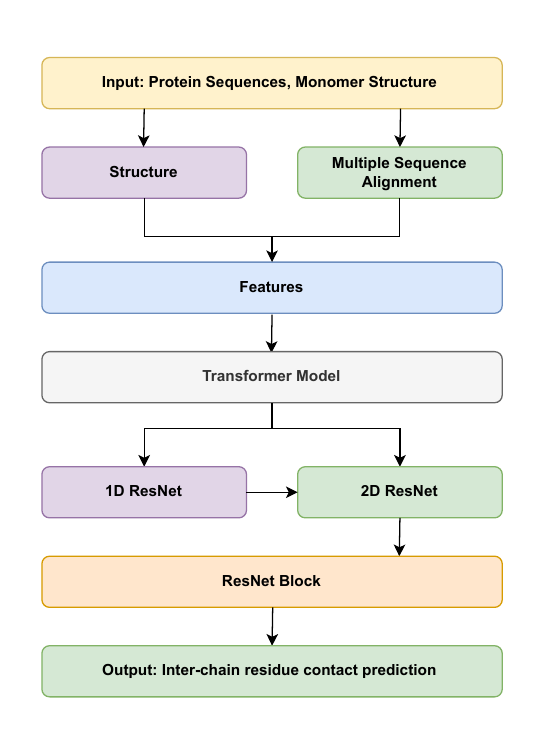}\hspace{2cm}
				\includegraphics[height=3.0in,width=3.0in]{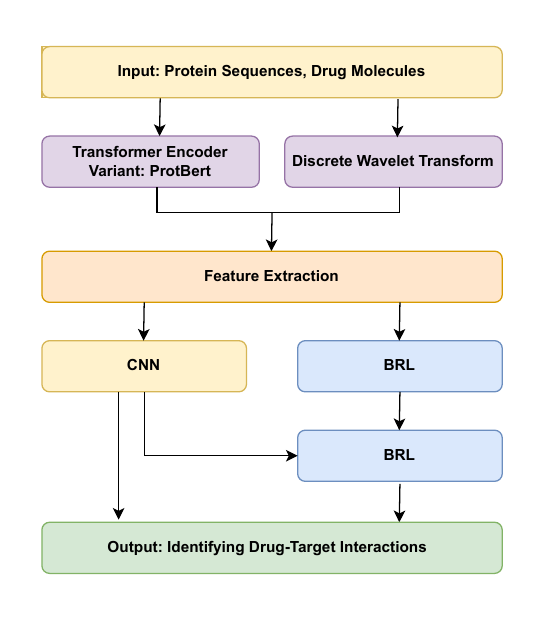}
                    }
				\centerline{(e)\hspace{95mm}(f)}
				\caption{Applications of Transformer-based language models to protein sequences for (a) OntoProtein, (b) LMPhosSite, (c) DeepZF, (d) DistilProtBert, (e) DeepHomo2 and (f) DTI-BERT}
				\label{aa}
			\end{figure*}

Gene Ontology (GO) is a systematic and comprehensive collection of ontologies related to gene attributes divided into  three non-overlapping domains of molecular biology: molecular function, biological process and cellular location. The science of protein is indispensable in understanding the disease therapies and human health \cite{vig2021}. In \cite{zhang2022ontoprotein}, the authors have used gene ontology (GO) embedding for protein pre-training which is the first framework of its kind to merge external knowledge graphs and protein pre-training (Figure \ref{aa}(a)). They propose a hybrid Encoder named OntoProtein which represents language text and protein sequence as well as use contrastive learning with knowledge-aware negative sampling to optimise knowledge graph and the protein sequence embedding while pre-training. To fulfill knowledge embedding, the GO notations are encoded as corresponding entity embeddings which are then optimised using simple knowledge embedding approaches \cite{Bordes2013}. They further leveraged the GO of molecular function, cellular component and biological processes. However, OntoProtein relies on GO annotations which may not be available for all proteins leading to poor embeddings of such proteins without GO labels. 
Another work~\cite{Vu2022} have proposed two models to predict GO using protein features as extracted by ProtBert~\cite{Elnaggar2022} model. Furthermore, they have customised and finetuned the ProteinBERT model to predict GO terms. Their results showed that protein embeddings that are created with such Transformer models can be used as a data source for tasks with sequence prediction as well as protein functions. Although the models are promising, this work suffers from the same drawback as \cite{zhang2022ontoprotein}.
Zhao et al. \cite{Zhao2023bio} have used a multi-view graph convolutional network model to extricate various GO representations from functional information, topological structure and their combinations. This model has used an attention mechanism to learn about the final knowledge representation of GO. Also, a pretrained language model viz. ESM-1b has been used to learn biological features for each protein sequence. The authors have tested the model on Yeast, Human and Arabidopsis and it has outperformed other state-of-the-art methods.
When compared with the SOTA methods, the model shows an improvement of 5\%, 4\% and around 2\% in AUPR for Yeast, Human and Arabidopsis respectively for Biological Process (BP). For Molecular Function (MF), such increase are 5\%, 3\% and 1\% respectively while for Cellular Component (CC), the improvements are around 3\%, 4\% and 2\% respectively. 
All the aforementioned works show potential but they do not focus on domain-invariant features, which may allow them to focus on function-relevant patterns and avoid dataset-specific biases. To mitigate this, a generalized approach, Graph Adversarial Learning with Alignment (GALA) is proposed in~\cite{Yfu2024} where the authors use adversarial training to learn domain-invariant features. Subsequently, they align the predicted protein representation to its GO term embedding, allowing the model to focus on function similarity rather than sequence similarity, resulting in good performance even for proteins with low sequence similarity.	Leveraging both sequences and structural information, the model provides biologically richer and more accurate feature representations. To showcase the efficacy of GALA, the authors have derived two versions of the same, viz. GALA-PDB and GALA. GALA-PDB is trained with a subset of proteins based solely on PDBch training set while GALA uses AlphaFold2-predicted protein structures during training. Both the models show impressive results with GALA scoring higher on most accounts. As compared to the other methods, GALA shows an improvement of 19\%, 3\% and 11\% for BP, MF and CC respectively in terms of AUPR. However, since this model uses a combination of GCN-based encoders, a multi-head meta-nodes graph Transformer, 2-layer MLP and finally a domain adversarial discriminator, the complexity of the GALA model and the high computational requirements limit its reproducibility and practical deployment. It is worth mentioning here that all the aforementioned works, although rely on similar concept, have used different datasets and thus comparing them in terms of results is is an unfeasible task in the framework of this survey.

\subsection{Functional and Structural Protein Cluster Identification}

Machine learning applied to amino acid sequences provides a solid and reliable way to predict functional and structural classification of proteins. Due to the extensive body of research in this area, we highlight a few key representative studies encompassing classification of membrane protein, prediction of phosphorylation site, classifying zinc-finger (ZF) motif as DNA-binding, prediction of glycolated protein, subcellular localisation prediction, adaptor protein classification, Enzyme prediction, peptide detection, predicting protein secondary structure and solubility, thermal stability and variants of certain significance (VUS). 

\textbf{Membrane Protein Classification}: Computational tools are applied for studying ion channels as well as membrane proteins and their corresponding functions. In this regard, \cite{Ghazikhani2022} has proposed TooT-BERT-M based on BERT model to detect membrane proteins and used logistic regression to classify membrane and non-membrane proteins. In this regard, they have the finetuned  ProtBERT-BFD model (viz. MembraneBERT) with membrane protein data to extract the embeddings and then and applied logistic regression for the final classification and named the final model as TooT-BERT-M. Evaluated on multiple benchmark datasets, TooT-BERT-M demonstrates high accuracy and MCC (92.46\% and 0.85) establishing the effectiveness of Transformer-based embeddings for membrane protein classification.
In \cite{Ghazikhani2023}, the authors have proposed TooT-BERT-C which uses BERT contextual representation to assess and discriminate ion channels from membrane proteins. This approach leverages finetuning of pretrained models such as ProtBERT‑BFD and MembraneBERT to generate contextual embeddings for each sequence and subsequently applies logistic regression for the final classification. Such approach shows that ProtBERT-BFD with logistic regression (TooT-BERT-C) performs better than MembraneBERT with logistic regression. On independent test set results, TooT-BERT-C shows an improvement of 1\% accuracy and 12\% MCC over MembraneBERT. When compared with SOTA, TooT-BERT-C reports better accuracy on both independent and cross-validation test sets (98.24\% and 98.96\%) while in terms of MCC it shows better results (0.85) for independent test set. While the results are promising, both TooT-BERT-M and TooT-BERT-C suffers from lack of interpretability. Without attention visualization or attribution analysis, the predictions remain a ``black box”, thereby limiting the practicality of the works.

\textbf{Phosphorylation Site Prediction}: Phosphorylation is yet another modification which plays an important role in various cellular processes. In \cite{Pakhrin2023}, Pakhrin et al. have proposed a deep-learning based method named LMPhosSite (Figure \ref{aa}(b)) for phosphorylation site prediction. The model enhances prediction performance by integrating embeddings from the local window sequence and the contextualized embedding generated using the global (overall) protein sequence from a pretrained protein language model. In order to capture effective local representation and global per-residue contextualized embedding from a pretrained protein language model. The LMPhosSite consists of two separate base models. A strategy called score-level fusion is used to integrate the output of the base models. For the combined serine and threonine independent test data set and the tyrosine independent test data set, respectively, LMPhosSite achieves precision, recall, Matthew's correlation coefficient, and F1-score of 38.78\%, 67.12\%, 0.390, and 49.15\%, which are higher than the compared approaches. Although LMPhosSite uses context of a full protein sequence, it is limited to residue-level embeddings, thereby leaving scope for predictive performance improvement. To mitigate this, Pratyush et al.~\cite{Pratyush2025} have proposed CaLMPhosKAN which utilises codon-aware embeddings from the codon adaptation language model (CaLM) to capture important information content inherent in the codon space. They combine the codon-aware embeddings with amino-acid embeddings from ProtTrans to create a bimodal representation of an input sequence. The prediction is based on fused embeddings analyzed via a Conv-BiGRU followed by a wavelet-based Kolmogorov–Arnold Network (WAV-KAN). The model achieves higher performance than existing predictors on both Ser/Thr (S/T) and Tyr (Y) test sets. It demonstrates robust generalisation, including in disordered protein regions. When compared with LMPhosSite on independent test set, CaLMPhosKAN shows an improvement in MCC (5\% and 7\% for S/T and Y residues respectively), Precision (34\% and 24\% for S/T and Y residues respectively) and AUPR (70\% and 50\% for S/T and Y residues respectively). With such great results, the model still suffers from some setbacks. The model requires alignment of protein and codon sequences, which adds computational and implementation overhead. Also, the fusion of two embedding types (codon-level + amino acid-level) increases model complexity and training cost. Moreover, the performance of the model is not explored on multi-site phosphorylation or under class imbalance in real proteomic datasets.

\textbf{Zinc-finger binding classifier}: The largest class of human transcription factors are Cys2His2 zinc-finger (C2H2-ZF) proteins which are essential for the regulation of gene expression and cellular function. In this regard, Gazit et al. \cite{Gazit2022} have proposed DeepZF (Figure \ref{aa}(c)) based on a protein Transformer for the prediction of binding ZFs and their DNA binding preferences by providing only the amino acid sequence of a C2H2-ZF protein. DeepZF is divided into two parts: BindZFpredictor which outputs the probability of the input ZF to bind DNA and PWMpredictor which provides the output position weight matrix for a given ZF sequence. The combination of these parts result in the final output of DNA-ZF binding preferences. The model has achieved an AUROC of 0.71 and a Pearson correlation of 0.42 in motif similarity. The work has also provided model interpretation to understand its inner working. Though the work is quite promising, it can be further improved by including independent microarray-based and sequencing-based datasets. Moreover, adding neighbouring ZFs may also help in the improved performance of the model. 

\textbf{Glycolated Protein Prediction}: Protein N-linked glycosylation is also an important post-translational mechanism in humans which contributes to vital biological processes. It usually occurs at the N-X-[S/T] sequon in amino acid sequence where X is any amino acid but proline. However, N-X-[S/T] sequon is a necessary but not sufficient determinant for protein glycosylation as not all N-X-[S/T] sequons are glycosylated. Thus, computational prediction of N-linked glycosylation sites confined to N-X-[S/T] sequons is an important problem. In this regard, Pakhrin et al. \cite{Pakhrin2023_1} have developed LMNglyPred (Language Model based N-linked glycosylation site Predictor) that uses embedding from ProtT5~\cite{Elnaggar2022} to predict N-linked glycosylation sites. The results have shown an accuracy of 86.93\% and an MCC of 0.717 on N-GlyDE’s independent test set. Although the model predicts whether a site is glycosylated, it does not offer insight into either the functional consequence of glycosylation or the conservation of sites across orthologs. Moreover, it shows a very high False Positive Rate and also there is a likelihood of the presence of some noise in the N-GlycoSiteAtlas dataset. Thus, the model needs better validated datasets for better prediction.

\textbf{Subcellular Localisation Prediction}: Subcellular localisation prediction is also very important for proteomics research. In this regard, Thumuluri et al. \cite{Thumuluri2022} have proposed DeepLoc 2.0, an update to DeepLoc \cite{Armenteros2017} (this model uses a recurrent neural network along with an attention mechanism) by adding multi-localisation prediction thereby improving both performance and interpretability. DeepLoc 2.0 uses a pre-trained protein Transformer language
model for feature representation while an attention plot is used to visualise which part of the input the model uses for its predictions. Finally, the prediction is performed for ten class subcellular localisation and nine class sorting signal tasks. The authors have also developed a web server, thereby making the model easily accessible to biologists who may not have much coding expertise. DeepLoc 2.0 uses embedding from ESM-2, ESM-1b and ProtT5 models and then applies multi-layer perceptron for the final prediction. Both ESM-1b and ProtT5 based models perform better than DeepLoc, providing an improvement in AUC of more than 6\% for different subcellular location prediction. Although, the model shows remarkable performance, it suffers from a drawback. The model only predicts among predefined subcellular locations (for example, cytoplasm, nucleus, mitochondria, etc.) and organism-specific locations are not accounted for.

\textbf{Adaptor Protein Classification}: In another important work, Rahardja et al. \cite{Rahardja2022} have used Transformer networks for the classification of adaptor proteins. Such proteins are crucial for intercellular signal transduction and their dysfunctionality may lead to several diseases. Keeping this in context, \cite{Rahardja2022} have proposed a novel Transformer based model encompassing a convolutional block and fully connected layer for protein classification. To perform such a classification, the authors extract PSSM features from the input protein sequences and subsequently process it using the proposed model. The experimental results show that the model has achieved an AUC of 0.903 and MCC of 0.487 on independent test dataset, thereby triumphing over the existing state-of-the-art methods. It is important to note that the paper highlights the model’s small size but does not provide clear inference of speed benchmarks or real-world deployment scenarios.

\textbf{Enzyme Prediction}: DAttProt have been proposed in \cite{Lin2022_1} to predict enzyme proteins. In this regard, Transformer Encoders have been used to encode sequences in pre-training while during finetuning local features are gathered by multi-scale convolutions. They have also designed a probabilistic double-scale attention weight matrix to combine multi-scale features and positional prediction scores. Finally, a full connection linear classifier performs a final inference by combining features and prediction scores. The model has been seen to perform quite well on DEEPre and ECPred datasets showing an accuracy of 0.788 on level 2 of DEEPre and macro-F1 of 0.967 on level 1 of ECPred. Here, level 1 represents one of the 6 main enzyme classes while levels 2 and 3 refer to subclass and sub-subclass respectively. With so many advantages, the model also suffers from the problem of random chopping off of longer protein sequences to a fixed length and thereby losing information in the process. 

\textbf{Peptide Detection}: A key step in protein identification and analysis is peptide detectability which is the probability of identifying a peptide from a mixture of standard samples. Such detection has been performed by \cite{Wang2022_1} using an ensemble deep learning model PD-BertEDL. The authors have used BERT to capture the context information of peptides and different deep learning techniques to represent the high quality features of different peptide categories. Subsequently, they use the average fusion strategy to combine the results from these three predictions to increase the flexibility of PD-BertEDL which shows an AUC value of 0.8832 and an AUPR value of 0.8401. Although, the model shows impressive results, it suffers from high computational complexity. 

\textbf{Protein Secondary Structure Prediction}: In order to interpret implicit structural and evolutionary information from three sequence-level classification tasks for protein family, superfamily, and fold, An et al. \cite{An2022} have developed a multi-task learning (MTL) architecture based on BERT in their study which includes MT-BERT (standalone BERT), MT-BCNN (BERT + CNN) and MT-BLSTM (BERT+BiLSTM). The knowledge as obtained in the MTL stage can also be transferred for downstream tasks of TAPE (Tasks assessing protein embeddings) which offers established standards to assess the performance of learned protein embeddings. To summarise, An et al. have used multi-task learning framework using three BERT-based backbones for contextual representations in natural language and also gathered knowledge on interrelated protein tasks. Finally, to assess whether the proposed MTL architectures effectively reflect the structural and evolutionary links, three downstream tasks with clear definitions in TAPE have been used; Secondary Structure  prediction (SSP), Contact prediction (CP) and remote homology (RH) detection. MT‑BCNN and MT-BLSTM show notably better precision in TAPE tasks; 0.82 vs 0.73 (MT‑BCNN), 0.45 vs 0.36 and 0.42  vs. 0.21 (MT-BLSTM) for SSP, CP and RH detection respectively. In another seminal work \cite{Wu2022}, the authors have proposed SPRoBERTa, a pre-training model based on RoBERTa \cite{liu2019} training framework, to explicitly model the local fragment of the protein sequence using an unsupervised protein tokenizer as learned from the data. Instead of using \textit{k}-mers to represent local fragments in protein sequences, the authors have used SentencePiece \cite{Kudo2018}, a subword encoding algorithm to tokenize the protein sequences. The authors have evaluated the effectiveness of SPRoBERTa by pre-training the Transformer \cite{NIPS2017} Encoder model on Pfam dataset and eventually finetuning the model on different tasks which encompass amino acid-level prediction, amino acid pair-level prediction and protein sequence-level prediction. On all the experiments, the pre-trained and finetuned models show significant increase in task performance, thereby showcasing the benefit of introducing local protein patterns into protein representation models. On TAPE tasks such as SSP, CH and RH, SPRoBERTa reports precision of 0.818, 0.632 and 0.304 respectively. Thus, both \cite{An2022} and \cite{Wu2022} have been able to surpass benchmark TAPE model. All the aforementioned works use full-scale Transformer models which are usually very resource intensive. To mitigate this problem, Geffen et al. \cite{Geffen2022} have employed a distilled model based on the concept of student and teacher networks for protein sequence analysis (Figure \ref{aa}(d)). The model, named DistilProtBert, is a distilled version of the ProtBert \cite{Elnaggar2022} model and can also be used to distinguish between real and random proteins. By using DistilProtBert, the authors have been able to reduce the size of the network and the running time by 50\% as well as a reduction of 98\% in computational resources for pretraining as compared to ProtBert model. While ProtBert was pretrained on around 216 M protein sequences, DistilProtBert uses around 43 M sequences during pretraining. To compare the finetuning results of both the models, downstream tasks such as SSP and membrane bound versus water soluble prediction have been carried out. DistilProtBert shows competitive accuracy for all the benchmark tasks; 0.75 (ProtBert) vs. 0.72 for CASP12, 0.83 (ProtBert) vs 0.81 for TS115 and 0.81 (ProtBert) vs. 0.79 for CB513 for secondary structure prediction. For membrane bound versus water soluble prediction, DistillProtBert shows an accuracy of 0.86 as opposed to 0.89 for ProtBert. This shows that although DistilProtBert is a much smaller model than ProtBert, it shows results on-par with ProtBert. On the other hand, DistilProtBert again shows good results for distinguishing real proteins from their shuffled counterparts. For singlet, doublet and triplet human proteome test sets, DistilProtBert shows an AUC of 0.92, 0.91 and 0.87 respectively while ProtBert has an AUC of 0.93, 0.92 and 0.87 respectively. Though, DistilProtBert shows remarkable results, there is no evidence yet for its effectiveness in more diverse protein-related tasks like structure prediction, function annotation or interaction modelling.  Also, when distilling from a single teacher model, student models like DistilProtBert struggle to fully match teacher performance, especially on complex tasks. This suggests a need for multi-teacher distillation strategies to close the performance gap. Chen et al.~\cite{CHEN2024} have proposed MFTrans that combines Multiple Sequence Alignment (MSA) Transformer~\cite{Rao2021} and multiple neural network components thereby offering a hybrid approach that allows the model to learn both short-range patterns and long-range ones. The model relies on multiple input features: multiple sequence alignment, position-specific protein matrix, hidden markov model profiles and Word2Vec. Combining such different features enables the model to learn richer representations and be more accurate in the prediction of protein secondary structure. It outperforms the existing benchmarks models outperforming their accuracy by 3\% on average. However, it relies on features derived from homologous sequences, thus the predictive performances drop for novel proteins or poorly conserved regions due to the lack of features of multiple sequence alignment. Table~\ref{tab:2_1} summarises the results of the above mentioned works for SSP (as they are all evaluated on CB513 benchmark dataset based on precision).
\begin{table}[H]
\centering
\begin{tabular}{|p{5cm}|p{2cm}|}
\hline
Model & Accuracy\\\hline 
TAPE~\cite{rao2019} & 0.73 \\\hline 
ProtBert~~\cite{Elnaggar2022} & 0.81\\\hline 
MTL~\cite{An2022} & 0.82\\\hline 
SPRoBERTa~\cite{Wu2022} & 0.81\\\hline 
DistilProtBert~\cite{Geffen2022} & 0.79\\\hline
MFTrans~\cite{CHEN2024} & 0.88\\\hline
\end{tabular}
\caption{Comparative analysis across models evaluated on CB513 benchmark dataset}
\label{tab:2_1}
\end{table}

Erckert et al. \cite{Erckert2024} have investigated whether addition of evolutionary information obtained from multiple sequence alignments (MSAs) with pLM can improve protein structure prediction. They probed into whether pLMs capture evolutionary information or pLMs and MSAs are correlated because they capture similar constraints on the protein sequences. They found that MSA coupled with older models likw SeqVec16 and ProtBert significantly improved the Secondary Structure prediction (SSP) performance as compared to modern pLMs like ProtT5. Moreover, inclusion of MSA degraded performance of pLMs in predicting intrinsically disordered regions of proteins which may be attributed to the difficulty of correctly aligning intrinsically disordered proteins. Thus, it can be interpreted that older models lacked sufficient capture of evolutionary patterns, hence benefit from explicit MSAs. Modern pLMs, on the other hand, already inherently encode evolutionary constraints, making external MSAs redundant, or sometimes even harmful. Further, MSA augmented bindEmbed21DL did not significantly enhance prediction of which residues in a protein bind to non-protein substrates. Tasks at three different levels viz. per residue, per protein and per segment level were compared using the same approach. Although their study showed that advanced pLM embeddings have relation with evolutionary information acquired from MSAs, they could not conclude whether there is only a correlation reflecting biophysical and functional constraints on protein sequences, or the same is explicitly/directly captured in the embeddings. This study thus reveals that modern pLMs, especially ProtT5, already encapsulate evolutionary information effectively, making the explicit integration of MSAs increasingly unnecessary and sometimes counterproductive. Evolutionary augmentation may still benefit legacy embeddings, but the state of the art lies in pure embedding-based prediction. Additionally, there is scope for investigation using latest pLMs along with other methodologies to integrate evolutionary information.

In~\cite{Heinzinger2024}, the authors present ProstT5 which is a protein language model capable of performing bidirectional translation between amino acids sequence and 3Di structural sequences (3D structural information encoded as 1D strings of tokens). The model is finetuned from Prot5 and allows to translate from sequence to structure and vice versa. ProtT5 used a simplified representation of structures, the 3Di alpahabet, which encodes the local environment of residues as discrete tokens. Unlike AlphaFold, which predicts 3D coordinates (x,y,z positions of every atom in a protein), this approach allows for significantly faster inference and scalability to large dataset. For the inverse folding task, the model generates amino acid sequences that can fold into a given 3Di structural environment. In doing that ProstT5 supports flexible sampling strategies that allows it to control variety and novelty. While techniques as top-k and top-p sampling limit the choice at the most plausible amino acid, the temperature scaling process allows to explore safer or newer sequences. For secondary structure prediction, ProstT5 shows an improvement in accuracy of about 4\% and 5\% as compared to one-hot encoding 3Di-sequences and ProtT5. The authors have constructed a highly non-redundant and diverse dataset composed solely of high-quality 3D protein structures, the authors aimed to maximize coverage of the sequence–structure space using as few proteins as possible. This strategy effectively minimized overrepresentation of large protein families commonly seen in both the PDB and sequence databases. However, this stringent filtering criteria may have introduced a different kind of bias. As a result, lProstT5 could potentially reflect and amplify these biases present in its training data. Another drawback might be the roundtrip translation between amino acid sequences and 3Di representation (AA - 3Di - AA, 3Di - AA - 3Di) that the model supports. Some discrepancies emerged from performing both full translational roundtrips indicating imperfect reversibility. This behaviour can lead to limitations in the model stability especially when implemented in iterative workflows, where repeated translations may accumulate errors and diverge from the original intent.

\textbf{Protein Solubility Prediction}: Wee et al. \cite{WEE2024} propose TopLapGBT which integrates pre-trained Transformer-based sequence embedding with persistent Laplacian (PL) features derived from 3D structures predicted by AlphaFold to predict protein solubility changes upon mutation. In this way TopLapGBT is able to capture both sequence (Transformers) and topological patterns. Although the model shows promising results (it improves upon the SOTA by 15\%), it does not offer an estimation of the prediction confidence.  

\textbf{Thermal Stability Predictions}: PTSP-BERT proposed in~\cite{LV2025} predicts the thermal stability of proteins using sequence-based bidirectional representations from Transformer-embedded features. The models used BERT-bfd for feature extraction and subsequently used six different classifiers, KNN, LGBM, RF, SVM, GNB and LR, to cross-validate the five sets of features obtained. When compared with the SOTA methods, the model shows superior performance. For the task of  distinguishing thermophilic proteins from non-thermophilic proteins, the model achieves an accuracy of 93\% and an MCC of 0.85, which is an improvement of 6\% and 14\% over the best model. For the task of distinguishing mesophilic proteins from non-mesophilic proteins, PTSP-BERT shows an improvement of 9\% and 33\% for accuracy and MCC respectively.
It can be noted though that the exact reasons behind better performances using BERT-bfd and dissimilar effects of different oversampling and undersampling methods after feature extraction are unclear; the model lacks clear interpretability. 

\textbf{VUS Prediction}: Joshi et al.~\cite{Joshi2025} have used ESM-2 (evolutionary scale model) embeddings to capture both generic (genome wide approaches) and gene specific features to predict variants of uncertain significance (VUS) in Arylsulfatase A. The method integrates zero-shot log-odds scores from the general ESM‑2 backbone and family-specific ESM embeddings into a supervised model trained on ARSA variants, yielding competitive or superior AUROC performance for predicting ARSA VUS and notably outperforming generic pathogenicity predictors when applied to NAGLU variants (0.89 as compared to 0.86 of best predictor model). This approach also included attention analysis that highlighted active and binding site residues linked to pathogenicity, offering interpretable mechanistic insights. Importantly, it demonstrated that in genes with limited experimental data, leveraging variants from the protein family can act as an effective proxy training source, achieving improved variant prioritization in diagnostic settings. However, their work is limited by the fact that it only works for well researched genes and gene families. 

\subsection{Generating \textit{de novo} proteins}
Protein design as well as classification and generation tasks using attention based models trained on protein sequences have shown unprecedented success.  Simple, conventional techniques that make use of numerous sequence alignments of related proteins, such as ancestral sequence reconstruction \cite{Gumulya2018}, have shown to be capable of producing valuable proteins but are constrained in their use. To mitigate this, many statistical and machine learning techniques have been used. However, most of these methods, starting from direct coupling analysis techniques \cite{Russ2020} to generative adversarial networks \cite{Repecka2021}, train on a fixed protein family in order to detect coevolutionary signals found within a collection of homologous sequences.
In this regard,~\cite{Madani2020} and~\cite{nijkamp2023} have proposed ProGen and a suite of protein language models referred to as ProGen2 based on autoregressive Transformers that are trained on various sequencing datasets gathered from over a billion proteins from genomic, metagenomic and immunological repertoire databases, and scaled up to 6.4B parameters. The authors have proposed a number of models viz. ProGen2-small (151M), ProGen2-medium (764M), ProGen2-large (2.7B) and ProGen2-xlarge (6.4B). The study investigates how increasing model scale and leveraging varied data distributions improve performance; larger ProGen2 models achieve significantly lower perplexity on held-out evolutionary sequences, indicating better modelling of natural sequence distributions. Importantly, without any task-specific finetuning, ProGen2 can generate novel sequences that fold into structurally valid proteins and predict fitness-related metrics in a zero-shot manner, outperforming previous benchmarks. The authors also emphasize that data diversity is as crucial as raw scale, suggesting future protein language models should prioritize balanced datasets from different biological sources. For zero-shot fitness prediction, ProGen2 has shown competitive performance when compared to ESM-2 and other SOTA models. For Adeno-Associated Virus (AAV) capsid fitness, Green Fluorescent Protein (GFP) brightness and Chorismate Mutase (CM) activity, ProGen2-xlarge has AUC of 0.678, 0.841 and 0.638 respectively against 0.512, 0.609 and 0.688 for ESM-2 with 3B parameters. However, ProGen2 is a decoder-only model thereby exhibiting limitations for those tasks that can be performed by encoder-based models. xTrimoPGLM, proposed in \cite{Chen2023} addresses this problem by proposing an encoder-decoder model which improves upon protein understanding and generation of new protein sequences concurrently by using a pre-training framework with 100 billion parameters and 1 trillion training tokens. In contrast to generic Encoder-only and causal decode-only language models, xTrimoPGLM employs the General Language Model's (GLM) \cite{Du2022} structural foundation to take advantage of its bidirectional attention and auto-regressive blank filling purpose. Building on the generating capability previously contained within the GLM goal, they extend the Masked Language Model (MLM) objective to the bidirectional prefix region to improve the representation capability of xTrimoPGLM. The training stage of xTrimoPGLM encompassed over 1 trillion tokens processed on a cluster of 96 NVIDIA DGX-A100 (8×40G) GPU nodes between January 18 and June 30, 2023. Comprehensive experiments have been conducted to evaluate the effectiveness of xTrimoPGLM where it has been seen to outperform other state of the art techniques. Moreover, xTrimoPGLM excels at \textit{de novo} protein generation, producing realistic protein sequences and after supervised fine tuning it supports programmable sequence design, underlining its versatility in both comprehension and creative protein engineering. xTrimoPGLM shows lower perplexity than ProGen2-xlarge in evaluations on two distinct out-of-distribution datasets, indicating its advanced performance; 13.35 and 8.78 as compared to 14.3 and 9.75 for ProGen2-xlarge. Although, the model shows commendable performance, critical enhancements are needed for real-world drug design which include tailoring models for a variety of protein-related tasks, increasing the accuracy of protein structure predictions, and minimizing unrealistic or biologically implausible outputs during protein generation. Moreover, it does not consider multiple modalities for protein generation. Another encoder-decoder model proposed by Yin et al.~\cite{yin2025} viz. Combinatorial Functional Protein GENeration (CFP-GEN) generates \textit{de novo} proteins with desired functional properties by combining diffusion modelling with large-scale protein language models. The method introduces a combinatorial prompt-based generation strategy, where functional protein representations (for example, GO terms, InterPro (IPR) domains and Enzyme Commission (EC) numbers) are encoded into a functional vector space using a pretrained protein language model DPLM~\cite{XWang2024}. These functional prompts are then used to guide a diffusion-based generative model (specifically, a denoising diffusion probabilistic model) to produce novel protein sequences that satisfy multiple target functions simultaneously. CFP-GEN innovatively fuses the compositionality of function (for example, merging two unrelated functional motifs) with sequence-level diversity, addressing challenges in existing generative models that often fail to preserve function when optimising for novelty. An essential goal in \textit{de novo} protein design is creating sequences that are both novel and diverse. To test CFP-GEN's ability to generate truly new proteins, the authors selected seven representative EC number classes (for example, Oxidoreductases, Transferases etc.). For each class, 30 sequences were generated and compared with real enzymes. Novelty was assessed by sequence dissimilarity to the closest training sequence, and diversity by differences from the full training set. CFP-GEN consistently showed higher novelty, indicating its effectiveness in producing original protein sequences rather than mimicking known ones. The model significantly outperforms baselines like ProGen2 (AUC of 0.795 vs. 0.663, 0.992 vs. 0.772, 0.951 vs. 0.661 for evaluating protein functionality via predicted GO terms, IPR domains and EC numbers respectively) in multi-objective protein design tasks, paving the way for efficient \textit{de novo} functional protein generation guided by desired biological properties. While CFP-GEN marks an initial effort to integrate multiple modalities for conditional protein generation, it comes with several limitations. Firstly, it currently accommodates only a limited range of GO, IPR, and EC annotations, which restricts its ability to generalize across a wider spectrum of functional categories. Secondly, for applications in bio-manufacturing, more comprehensive conditional inputs—such as physicochemical properties like hydrophobicity, charge, and polarity—are crucial but currently unsupported. Additionally, incorporating sequence–structure co-design is necessary to enable a fully end-to-end protein design framework.

Ferruz et al. \cite{Ferruz2022} have proposed ProtGPT2 with 738 million parameters, an autoregressive Transformer based language model that has been trained with 50 non-annotated million parameters spanning the protein space and produces \textit{de novo} protein sequences by mimicking natural ones. While disorder predictions show that 88\% of ProtGPT2-produced proteins are globular, in line with natural sequences, the created proteins exhibit natural amino acid propensities. Furthermore, sequence searches in protein databases show that ProtGPT2 sequences are quite related to natural ones as well as ProtGPT2 samples unexplored regions of protein space. Unlike ProGen2, ProtGPT2 cannot condition generation on function or family; it follows unconditional sampling. A key drawback of ProtGPT2 is its reliance on sequence-level training without explicit structural or functional supervision, which means that while it can generate proteins that statistically resemble natural ones, there is no guarantee that the generated sequences will fold correctly or perform any desired biological function. While ProtGPT2 focuses on purely generative modelling of protein sequences, a very important multitask model that performs both sequence generation and regression is Regression Transformer (RT) which is proposed by \cite{Born2023}. It uses regression as a conditional sequence modelling problem and can be trained on both numerical and textual tokens. The authors have derived a permutation language modelling (PLM) (an extension of masked-language modelling to autoregressive models) which is an alternate training scheme with a novel self consistency loss for better text generation using continuous primers. In this work, the authors have demonstrated the capabilities of RT on various set of predictive and generative tasks in chemical and protein language modelling. Unlike traditional language models that are either generative or predictive, RT integrates regression heads directly into the Transformer decoder, enabling it to generate valid molecular SMILES (Simplified Molecular Input Line Entry System) strings while concurrently predicting associated continuous-valued molecular properties such as solubility, logP, and binding affinity. This dual-capability model is trained end-to-end using a masked language modelling objective for generation and a mean squared error (MSE) loss for regression, effectively learning joint representations. The model outperforms existing baselines in both property prediction accuracy and conditional molecule generation, demonstrating strong generalization, especially in low-data regimes. Additionally, RT supports property-conditioned molecule generation, where molecules can be designed to meet target properties, making it highly applicable for drug discovery and materials science tasks.

In their work, Dumortier et al. \cite{Dumortier2022} have proposed PeTriBERT, proteins embedded in tridimensional representation in a BERT model to predict an amino acid primary sequence from protein 3D structure. PeTriBERT inverse folding model is a simple Transformer model augmented with 3D-structural data which has been trained on more than 350,000 protein sequences as retrieved from AlphaFoldDB database \cite{Varadi2021} and is used \textit{in silico} to generate completely new proteins having GFP-like structure. The novelty of the proteins has been confirmed as 9 out of 10 of these GFP structural homologues have not been found in proteome database, thereby making PeTriBERT a valuable tool for new protein design. The primary distinction between PeTriBERT and pure NLP models is the incorporation of additional embedding modules, 3D position embedding and rotation embedding to provide the model with tridimensional data. The model is trained to predict the amino-acid sequence in a traditional masked language modelling setup. Despite these remarkable achievements, PeTriBERT suffers from the training on AlphaFold-predicted backbones, which are not experimentally validated. Any systematic biases or errors in these predicted structures (for example, imprecise sidechain packing or flexible loops) may propagate into the model, affecting the quality of designed sequences. While the model has shown success in single-case \textit{de novo} designs for GFP-like proteins, broader benchmarking, such as functional assays across different folds or detailed stability profiling, is still lacking, leaving open questions about its robustness in practical protein engineering scenarios.

In~\cite{Hayes2025}, Hayes et al. have proposed ESM-3 model which is a major milestone in protein language modelling. With 98 billion parameters and trained on 2.78 billion protein sequences, ESM‑3 is one of the largest protein LMs to date. It features a multi-track architecture that simultaneously reasons over sequence, structure, and function, using a masked language modelling objective. Remarkably, ESM‑3 can generate entirely new proteins; for example, researchers used it to design ``esmGFP” which a novel green fluorescent protein with just ~58\% sequence identity to known examples making it equivalent to simulating an evolutionary distance of around 500 million years. ESM-3 has extended protein modelling beyond monomeric structure prediction (as in ESM‑2 or AlphaFold) by enabling multimodal generative design. It also demonstrates evolutionary ``chain-of-thought” generation, making it a powerful tool for novel protein design and synthetic biology. Along with all these advantages, ESM-3 suffers from the major drawback of requiring high computational resources for execution making it difficult for finetuning the model. While ESM-3 supports structural reasoning, its monomer structure prediction still lags behind AlphaFold2 in accuracy, particularly for challenging or disordered proteins.

\subsection{Binding}
Binding between molecules is a thermodynamically driven process, which in the case of proteins can be investigated through primary sequences by applying a machine learning approach. Protein-protein interactions have a great impact in several biological processes. The release of AlphaFold~\cite{Senior2020} and its more refined successor, AlphaFold2~\cite{jumper2021}, marked a major leap in monomeric protein structure prediction.  However, in practical applications, accurate structure prediction was still challenging, particularly for proteins with few or no homologous sequences, since AlphaFold2 relies heavily on evolutionary information derived from multiple sequence alignments.
Thus, in order to predict inter-protein connections of homo-oligomers, a specialized model was created in the form of DeepHomo2.0~\cite{Lin2022} (Figure \ref{aa}(e)). It is a Transformer based learning model that utilises the direct-coupling analysis (DCA), Transformer characteristics of sequences, and the structural properties of monomers to predict protein-protein interactions of homodimeric complexes. Using experimental monomer structures, DeepHomo2.0 achieves $>$70\% precision among the top 10 contacts, but this drops to ~60\% when using predicted monomer structures. Thus, its accuracy depends on high-quality monomer models. Errors in those structures directly affect contact prediction quality. To address the need for quaternary structure prediction at scale, AlphaFold-Multimer~\cite{Evans2021} extended AlphaFold2 by incorporating inter-chain co-evolutionary signals to model multimeric protein complexes. More recently, AlphaFold3~\cite{Abramson2024} has further advanced this frontier by accurately predicting the structure of proteins and ligands and how they interact.
On the other hand, Drug-Target interactions (DTIs) is another aspect which is very important for genomic drug discovery as well as finding a potential drug for a target. Zheng et al. \cite{Zheng2022} have proposed DTI-BERT (Figure \ref{aa}(f)) that uses pre-trained ProtBert to extract sequence features from target proteins. The authors have also used other state-of-art drug descriptions like Word2vec, Node2vec, Graph Convolutional Network (GCN) and Discrete Wavelet Transform (DWT) where DWT has shown the best performance. Initially, DWT has been used to generate information from drug molecular fingerprints. Subsequently, the feature vectors of DTIs have been concatenated and provided as inputs to feature extraction module made up of a batch-norm layer, rectified linear activation layer and linear layer, called BRL block and a convolutional neural networks module to extract the features of DTIs. Finally, a BRL block has been used as the prediction module. The prediction accuracy of DTI-BERT for various target families of ion channels, nuclear receptors, G-protein-coupled receptors and enzymes are respectively 94.7\%, 89\%, 90.1\% and 94.9\%, thereby indicating the superiority of DTI-BERT over other existing methods.  

Experimental analysis of molecular recognition features (MoRF) which is significant in protein interactions can be time consuming and costly. To mitigate this problem,~\cite{Meng2025} introduces Trans-MoRFs which consists of an embedding layer followed by the Transformer block (two encoders and two decoders) and finally a classification layer to predict whether each position of the sequence belongs to a disordered region or not. Accuracy and area under the curve are improved as evaluated on all seven datasets at a low computational cost compared to existing state of the art models. Further, protein sequences of varied lengths with long or short length MoRF regions can be handled by the model. With respect to AUC, Trans-MoRFs has been evaluated on multiple test datasets like TEST419 (0.93), TEST464 (0.93), TEST266 (0.94), TESTNEW (0.92), EXP53 (0.94), EXP53\_long (0.94) and EXP53\_short (0.94). However, Trans‑MoRFs, like earlier MoRF predictors, relies heavily on experimentally annotated MoRFs from PDB structures, with negative samples drawn from surrounding residues. However, this approach risks false negatives, as some ``non-MoRF” residues may actually be undiscovered binding motifs. This can skew the model’s decision boundaries and inflate its apparent performance. Also, to improve its prediction accuracy, integrating the structural information of proteins may be integrated while inclusion of cross-species data may improve generalizing capability of the model. One work which has addressed the cross-species generalisation is Ko et al.~\cite{Ko2024}. They have proposed TUnA to minimise false positive protein–protein pairs by lowering confidence for predictions involving unfamiliar protein pairs or out of distribution samples. TUna helps in saving time and money by identifying most certain PPI candidates for experimental validation along with capability of generalizing to proteins of new species outside the training set. TUna uses ESM-2 for embeddings followed by spectral-normalized Neural Gaussian process layer. Accuracy of TUnA on the human only dataset was improved to 0.62 from 0.56 as obtained by the state of the art Topsy-Turvy model. However, in TuNA inter protein shared weights use the concatenation form of $A||B$ and $B||A$, (where $A$ and $B$ are the protein sequences) leading to extra computations. Also, despite being an uncertainty-aware model, TUnA does not leverage unlabeled protein sequences to improve generalization.

Detecting linear B-cell epitopes (BCEs) is another important method which is essential for vaccine design, immunodiagnostic test, antibody production, disease prevention and treatment. In this regard, wet lab experiments are very expensive and labour-intensive, thereby paving the way for low-cost computational methods. Liu et al. \cite{Liu2023} have proposed LBCE-XGB to predict linear BCEs based on XGBoost algorithm. In order to reflect the biological information hidden in peptide sequences, the embeddings of the residues have been derived using a domain-specific pre-trained BERT model. Also, other attributes encompassing amino acid composition and antigenicity scale have also been considered and based on cross-validation results, the best feature combination was determined.
Furthermore, the best feature subset from the 768 dimensional vector as obtained from BERT model has been determined using Shapley Additive explanation (SHAP) algorithm. This led to the identification of top 200 features that have been selected to build the models. Finally, based on an AUROC value of 0.765, the top 130 features have been selected. Moreover, based on further AUROC value of 0.845, the feature group of amino acid composition (AAC), amino acid pair antigenicity scale (AAP), amino acid trimer antigenicity scale (AAT) and BERT have been selected as the optimal combination of features. All the conducted experiments show that LBCE-XGB is robust on both training and independent test data set. 
LBCE-XGB has been compared with several linear B cell epitope predictor and it achieves an AUROC of 0.845 on the training dataset which is 3.6\% higher than the SOTA while on the test dataset, the model achieves an AUROC of 0.838.
While the results are encouraging,  using SHAP after the embeddings may affect some of BERT's contextual richness for interpretability and simplicity. 

Another model that utilises the strength of Transformers is~\cite{Wu2024} where the model does not require structural or domain knowledge, but uses only the primary amino-acid sequence. The authors propose two models to generate binder sequences that interact with a given target protein. While AppendFormer works well for short prior and has a simple architecture, with MergeFormer they propose an enhanced version of the first, that encodes both the receptor sequence and binding score before merging them in the latent space. AppendFormer shows an accuracy of 95.81\% while MergeFormer has an accuracy of 96.41\%. While the models can generate sequences with high sequence similarity to known binders, there is no experimental evidence (for example, binding affinity assays, structural docking etc.) to confirm that the generated sequences actually bind the target proteins. This limits the biological relevance of the predictions. In contrast,  GeoDock introduced in~\cite{ChuRu2024} leverages 3D structural representations and multitrack iterative attention to model flexible docking between protein pairs, predicting how two proteins physically interact at the atomic level. This improvement is particularly relevant for modelling binding-induced conformational changes, making it more realistic and biologically relevant. This model uses the multitrack iterative Transformers architecture, similarly to AlphaFold2 but with a main difference, it multiple sequence alignment free. That makes the model faster, easier to use since it does not depend on external tools and is applicable even for proteins with no homologs. However, the model carries out the refinement of sidechains relying on RoseTTA introducing a dependency from an external tool. 

Jahn et al.~\cite{Jahn2024} have introduced IDBindT5 to predict binding residues in intrinsically disordered proteins or regions using embedding from pLMs (ProtT5). Unlike traditional predictors that rely on evolutionary features like MSAs or handcrafted biophysical descriptions, the model proposed by the authors leverages only single-sequence embeddings (vector representation of a protein generated from just the amino acid sequence without using multiple sequence alignment or any additional context annotation) and disordered annotations to achieve improved performance when compared to SOTA models. When tested on Mobi195 which has 195 proteins, IDBindT5 has a balanced accuracy of 57.2$\pm$3.6\% (95\% CI) against SOTA methods like ANCHOR2 (52.4 $\pm$ 2.7\%) and DeepDISOBind (56.9 $\pm$ 5.6\%). For  Critical Assessment of protein Intrinsic Disorder round 2 (CAID2) (78 proteins) benchmark data, IDBindT5 shows the same performance as DeepDISOBind and performs better thab ANCHOR2. The model is efficient and less resource-intensive where it is designed to run efficiently on consumer grade GPUs or even CPUs, being able to make residue-level binding prediction across an entire human proteome in a few hours. This computational efficiency is crucial in large-scale studies where thousands of proteins need to be processed in batch. Despite its simplicity, the model achieves SOTA or near-SOTA results on major benchmark datasets. The model either matches or outperforms methods like DeepDISOBind, which rely on more complex and resource-heavy input features, showing how LM embeddings are not just computationally convenient, but also biologically informative. The model outputs binding probabilities for each individual residue, rather than just a binary prediction at the protein level. This allows for more applications such as identifying specific binding regions or motifs within intrinsically disordered segments that are likely to be involved in binding, prioritizing residues for mutagenesis or guiding peptide design by selecting the most functionally relevant protein subsequence. While there are many strengths of the model, there are also some weaknesses worth discussing. The model does not specify the ligand-type; it predicts binary binding status but the knowledge of whether the binding is to the DNA, RNA, protein or small molecules is missing. This limits the model usefulness in functional proteomics or therapeutic design where knowing the binding partner is critical. Moreover, the authors have restricted their training and testing data to disordered regions only, making the model currently unable to predict binding residues in ordered proteins or in semi-disordered regions. By focusing solely on disordered regions, the model cannot generalise to ordered or hybrid proteins, potentially missing on relevant binding sites.

Table~\ref{tab:2} provides a comprehensive overview of the mentioned works for each of the concerned categories.

\onecolumn
\begin{landscape}
\scriptsize
\begin{center}
\begin{longtable}{p{2cm}p{1.5cm}p{3cm}p{3cm}p{3cm}p{3cm}p{4cm}p{2cm}}
\caption{Brief description of representative applications of Transformer-based language models for Protein sequences} 
\label{tab:2}\\\hline
Category  & Paper & Input & Output & Data Resource & Task Description & Major Results & Data Repository \\\hline
\endfirsthead
\hline
Category  & Paper & Input & Output & Data Resource & Task Description & Major Results & Data Repository \\\hline
\endhead
       Gene Ontology (GO) & Zhang et al. \cite{zhang2022ontoprotein} & Protein Sequences, Gene Ontology & GO-aware protein representations &  ProteinKG25, Swiss-Prot
       & Using knowledge graphs for protein pretraining   & Improves performance on protein–protein interaction (PPI) prediction and protein function prediction tasks & \href{https://github.com/zjunlp/OntoProtein}{OntoProtein} \\\cline{2-8} 
       & Vu et al. \cite{Vu2022} & Protein Sequences, GO Terms & Gene Ontology Functions for each protein & \url{http://purl.obolibrary.org/obo/go.obo}, UniProtKB/Swiss-Prot &  Using ProtBert to extract protein features and finetuning ProteinBERT to predict GO terms  & GO Annotations for Proteins & N/A\\\cline{2-8}
        & Zhao et al. \cite{Zhao2023bio} & Protein Sequences, GO Terms & Protein–GO term affiliation scores & Uniprot, \url{https://www.ebi.ac.uk/GOA/downloads}, \url{https://geneontology.github.io/docs/download-ontology}
        & Fusion of functional and topological knowledge of GO to guide 
        protein function prediction & Outperforms SOTA baselines in all three species benchmarks (Yeast, Human, Arabidopsis) & \href{https://github.com/Candyperfect/Master}{Protein Function Prediction}\\\cline{2-8}
       &  Fu et al. ~\cite{Yfu2024} & Protein Sequences and Structures, GO Terms & Multi-label protein function prediction, mapping proteins to GO & PDB-chains (PDBch), AlphaFold2-predicted chains (AFch), Swiss-Prot &
       Using adversarial training to learn domain-invariant features & Demonstrates state-of-the-art performance on function prediction benchmarks, achieves interpretability by using Grad-CAM~\cite{Selvaraju2020} & \href{https://github.com/fuyw-aisw/GALA}{GALA}\\\hline
        Functional and  structural protein & Ghazikhani et al. \cite{Ghazikhani2022} & Protein Sequences & Binary classification to predict whether proteins are membrane or non-membrane & \url{ https://tootsuite.encs.concordia.ca/datasets/membrane} & Detection of membrane proteins based on BERT and Logistic Regression & Compared frozen embeddings vs finetuned embeddings where finetuned models outperformed frozen representations & \href{https://github.com/bioinformatics-group/TooT-BERT-M}{TooT-BERT-M}\\\cline{2-8} 
          & Ghazikhani et al. \cite{Ghazikhani2023} & Protein Sequences & Binary classification to predict whether a protein is an ion channel versus a non-ion membrane protein & \url{http://bio216.bioinfo.yzu.edu.tw/deepion/app/download/} & Utilises the BERT contextual representation to assess cluster identification and discriminate ion channels from membrane proteins via a Logistic Regression classifier & Demonstrated state-of-the-art discrimination compared to prior methods  & \href{(https://huggingface.co/ghazikhanihamed/IonchannelBERT}{TooT-BERT-C}
       \\\cline{2-8} 
            &  Pakhrin et al. \cite{Pakhrin2023} &  Protein sequences & Binary classification to predict whether a residue is a phosphorylation site & Swiss-Prot, dbPTM, phosphoELM, PhosphoSitePLUS & Phosphorylation site prediction & Performance surpasses existing methods for both residue types (S/T and Y sites) &  \href{https://github.com/KCLabMTU/LMPhosSite}{LMPhosSite}\\\cline{2-8} 
       & Pratyush et al.~\cite{Pratyush2025} & Protein sequences and their underlying coding DNA sequences  & Binary classification of S/T and Y residues as phosphorylated or not & \url{https://covinet.innatelab.org}, \url{https://github.com/gankLei-X/DeepPSP}, \url{http://github.com/dukkakc/Chlamy-EnPhosSite} & Prediction of general protein phosphorylation sites using sequence-based features & Demonstrated higher predictive performance, including on residues in disordered regions & \href{https://github.com/KCLabMTU/CaLMPhosKAN}{CaLMPhosKAN} \\\cline{2-8}
        & Gazit et al. \cite{Gazit2022} & Human C2H2-ZF protein sequences & Classifies each zinc-finger (ZF) motif as DNA-binding or non-binding & ChIP-seq experiment, In vivo Binding-ZF-classification dataset, B1H dataset & A deep-learning-based pipeline for predicting binding ZFs and their DNA-binding preferences & Model interpretation revealed biologically meaningful sequence–function relationships, pinpointing critical residues in the ZF recognition helix & \href{github.com/OrensteinLab/DeepZF}{DeepZF}\\\cline{2-8} 
        & Pakhrin et al. \cite{Pakhrin2023_1}& Protein sequences & Classify each N‑X‑[S/T] sequon in human proteins as glycosylated or non-glycosylated & N-GlycositeAtlas and N-GlyDE & Predicting N-linked glycosylated sites in human proteins & Understanding where glycosylation occurs can be vital for various biological and medical research areas & \href{https://github.com/KCLabMTU/LMNglyPred}{LMNglyPred}\\\cline{2-8} 
        & Thumuluri et al. \cite{Thumuluri2022} & Protein sequences & Multi-label subcellular localisation across 10 cellular compartments, Sorting signal prediction for 9 signal types & UniProt & 
        Predicting protein subcellular localisation and signal classification & High accuracy, Attention-signal correlation provides better interpretability & \href{https://services.healthtech.dtu.dk/services/DeepLoc-2.0/}{DeepLoc-2.0}\\\cline{2-8} 
         & Rahardja et al. \cite{Rahardja2022} & Protein sequences & Binary classification of each protein as an adaptor protein or non-adaptor protein & UniProt, Gene Ontology & Protein classification & Confirmed model stability and generalizability using rigorous cross-validation & \href{https://github.com/wangmou21/adaptor}{Adaptor}\\\cline{2-8} 
        & Lin et al. \cite{Lin2022_1}& Protein sequences & Hierarchical enzyme class prediction & DEEPre and ECPred datasets & Enzyme vs. non-enzyme classification and predicting enzyme commission (EC) numbers at increasing specificity & High accuracy on the DEEPre dataset and high Macro-F1 score on ECPred dataset, double-scale attention mechanism effectively identifies biologically meaningful sub-sequences and functional motifs & \href{https://github.com/zhanglabNKU/DAttProt}{DAttProt}\\\cline{2-8}
       & Wang et al. \cite{Wang2022_1} & Peptide sequences & Prediction score (probability) of detecting a peptide in mass spectrometry-based proteomics & GPMDB database & Predicting peptide detectability & Shows enhanced robustness and adaptability due to the ensemble strategy & N/A\\\cline{2-8}

        & An et al. \cite{An2022} & Protein sequences & Sequence-level classification & SCOP2 classification dataset & Designing multi-task learning architecture to classify protein family, superfamily and fold& Outperforms baseline Transformer models on remote homology detection, contact prediction, and secondary structure tasks & \href{https://scop.mrc-lmb.cam.ac.uk/download}{Models}\\\cline{2-8} 
       & Wu et al. \cite{Wu2022} & Protein sequences & Pre-Trained model finetuned for secondary structure prediction (SSP), contact map prediction (CMP), remote homology detection (RHD) and protein function classification (PFC) & Pfam, UniRef50 (pre-training), NetSurfP-2.0/CB513 (SSP),  CASP12 (CMP), SCOP (RHD) &Understanding protein function and structure in computational biology & Improved accuracy in SSP, CMP and RHD while for PFC the model shows competitive performance & N/A\\\cline{2-8} 
        & Geffen et al. \cite{Geffen2022}& Protein sequences & Distilled PLM   & UniRef50 (Training), CASP12, TS115, CB513 (Secondary Structure Prediction), DeepLoc (Membrane protein)  &  finetuning tasks and distinguishing between real proteins and their randomly shuffled counterparts & Model compression where training runtime is halved and pretraining cost reduced by 98\%  & \href{https://github.com/yarongef/DistilProtBert}{DistilProtBert}\\\cline{2-8} 
        & Chen et al.~\cite{CHEN2024} & Protein sequences & Predicting protein secondary structure &  CASP12, CASP13, CASP14, TEST2016, TEST2018 and CB513 &Strong generalisation and mutation adaptability, especially at residue-level resolution & Demonstrated enhanced performance in predicting effects of mutations on secondary structure & N/A\\\cline{2-8} 
        & Erckert et al. \cite{Erckert2024} & Protein sequences & Multiple downstream protein prediction tasks  & PDB, ConSurf10k, CheZOD117, CASP12, CASP14, SignalP-5.0, ASTRAL 2.06, CAFA3 & Integration of evolutionary information from MSAs with embeddings from pLMs for subsequent supervised learning protein prediction tasks & For SeqVec and ProtBert, the combination significantly improved accuracy but ProtT5 did not benefit from adding MSAs &https://github.com/erckert/EV-embeddings \\\cline{2-8} 
        & Heinzinger et al.~\cite{Heinzinger2024} & Protein sequences and 3Di structures & Bidirectional sequence-structure translations & AlphaFold Protein Structure
        Database (AFDB) & Performs both forward (folding) and inverse (design)
        translation between amino acid sequence and 3Di structure using a
        bilingual LM & Fast, flexible and effective generation of novel sequences and structures; up to 1000 times faster than AlphaFold for 3Di & https://huggingface.co/Rostlab/ProstT5\\\cline{2-8} 
       & Wee et al.~\cite{WEE2024}& Protein sequences and structures & Predicting protein solubility changes upon mutation & Multiple experimental sources & Integrates Transformer-based sequence embeddings with topology-based structural features derived from persistent Laplacians & Outperforms prior solubility prediction tools like CamSol, PON-Sol2, and others on benchmark datasets & \href{ https://github.com/ExpectozJJ/TopLapGBT}{TopLapGBT}\\\cline{2-8} 
         &Lv et al.~\cite{LV2025}& Protein sequences & Protein thermal stability predictions & \href{https://github.com/ailanhuang/A-machine-learning-model-for-psychrophilic-proteins}{Huang et al.} and IND-Enzymes database & Predict protein thermal stability by effectively using the information encoded within protein sequences through BERT-based representation learning.  & Demonstrates effective thermal stability prediction using sequence-based Transformer embeddings, outperforming baselines relying on handcrafted features & \href{https://github.com/zhibinlv/PTSP-BERT}{PTSP-BERT}\\\cline{2-8} 
       &Joshi et al.~\cite{Joshi2025} & Protein sequences & Scores indicating likely impact (activity change) of each amino acid variant & UniProt, Clinvar,  gnomAD, Pompe disease mutation database and LOVD 3.0 database & Predicting variants of uncertain significance & Inclusion of family-level variant data significantly improved performance, even for genes with limited curated variant labels & N/A \\\hline

        Generating \textit{de novo} proteins & Madani et al.~\cite{Madani2020} & Protein sequences, taxonomy, keyword tags & Conditional protein sequence generation & Uniparc, UniprotKB, SWISS-PROT, TrEMBL, Pfam and NCBI & Designed to produce novel protein sequences with specified properties via control tags & 1.2B-parameter decoder-only Transformer, generalizes well to unseen protein families & \href{https://github.com/salesforce/progen}{Progen}\\\cline{2-8}
         & Nijkamp et al. \cite{nijkamp2023}& Protein sequences & Novel protein sequences & UniRef90, BFD metagenomic database & \textit{De novo} sequence generation, Zero-shot fitness prediction & Successful generation of viable, foldable proteins, strong performance in zero-shot fitness prediction & \href{https://github.com/salesforce/jaxformer}{Progen2}\\\cline{2-8} 
         & Chen et al. \cite{Chen2023}& Protein sequences & \textit{De novo} protein generation, structure-aware sequence design and 3D structure prediction & Large-scale unlabeled protein corpus (pretraining), \url{https://huggingface.co/proteinglm} (downstream tasks) & Generating novel protein sequences structurally similar to natural ones & Outperforms baselines on 18 protein understanding benchmarks (across structure, function, localisation, etc.), enables atomic-resolution structural predictions and supports \textit{de novo} protein generation & \href{https://github.com/ONERAI/xTrimoPGLM}{xTrimoPGLM}\\\cline{2-8} 
        & Yin et al.~\cite{yin2025} & Protein sequences, Functional Annotations (GO Term, IPR Domain and EC Number) & Novel protein sequences & UniProtKB, InterPro and CARE databases & To generate \textit{de novo} functional proteins conditioned on enzyme function & Generated proteins are highly novel (low sequence similarity to training set) & \href{t https://github.com/yinjunbo/cfpgen}{CFP-GEN}\\\cline{2-8}
         
      & Ferruz et al. \cite{Ferruz2022} & Protein sequences &  Novel protein sequences & UniRef50 & Generating \textit{de novo} protein sequences based on principles of natural ones & Natural-like properties of the generated sequences, structural novelty and samples unexplored regions of protein space & \href{https://huggingface.co/docs/Transformers/main_classes/trainer}{ProtGPT2}\\\cline{2-8}

        & Born et al. \cite{Born2023}& Protein and molecular sequences (SMILES) & Property prediction and novel protein sequence generation & Multiple sources \url{https://github.com/IBM/regression-transformer} & A multitask model for property prediction and conditional generation & Abstracts regression as a conditional sequence modelling problem and  excels at property-driven conditional generation & \href{https://github.com/GT4SD/gt4sd-core/tree/main/examples/regression_Transformer}{Regression Transformer}\\\cline{2-8} 
     
       &  Dumortier et al. \cite{Dumortier2022}& Protein sequences and 3D structures & Predicted amino acid sequences mapping to input backbone structures & AlphaFoldDB & Inverse protein folding/design & Generating \textit{in silico} new proteins & \href{https://github.com/Baldwin-disso/PeTriBox}{PeTriBERT}\\\cline{2-8}

         &Chennakesavalu et al.~\cite{Chennakesavalu2024} & Rotamer library & Ensembles of side-chain conformations, Conformational diversity metrics & Molecular Dynamics (MD) simulation & Generation of protein conformational ensembles & Atomistic accuracy in side-chain positioning and diverse conformational ensembles & \href{https://github.com/rotskoff-group/transformer-backmapping}{Backbone-to-side chain Transformers}\\\hline

       Binding & Lin et al. \cite{Lin2022}& Protein sequences, Structure features of monomers & Inter-chain residue–residue contact prediction in homodimeric proteins & \url{http://huanglab.phys.hust.edu.cn/DeepHomo/} & DeepHomo2.0 uses sequence + monomer structure, DeepHomoSeq uses only sequence & With experimental monomer structures: $>$70\% precision in top‑10 contacts, With predicted monomer structures (via AlphaFold2): $>$60\% precision, DeepHomoSeq (sequence-only): $>$55\% top‑10 precision
        & \href{http://huanglab.phys.hust.edu.cn/DeepHomo2/}{DeepHomo2}\\\cline{2-8} 
       & Zheng et al. \cite{Zheng2022} & Protein sequences, Drug molecules & Binary classification of drug–target interactions (DTIs) in cellular networks & \href{https://journals.plos.org/plosone/article?id=10.1371/journal.pone.0009603#pone.0009603.s001}{He et al.} & Identifying Drug-Target interactions  & High prediction accuracy for Ion channels (94.7\%), Enzymes (94.9\%), GPCRs: (90.1\%) and Nuclear receptors (89\%) & \href{https://github.com/Jane4747/DTI-BERT}{DTI-BERT}\\\cline{2-8} 
     
       &Meng et al.~\cite{Meng2025} & Protein sequences & Binary classification to predict
whether protein sequences are MoRF or non-MoRF residues & Multiple sources~\cite{Disfani2012,Malhis2015}& Predicting disordered protein &  Improved AUC on all benchmark datasets & N/A\\\cline{2-8}
   & Ko et al.~\cite{Ko2024} & Protein sequences & Binary PPI prediction, Uncertainty estimate between 0 and 1 & STRING database (v11) (Cross-Species), Bernett dataset & Predicting unfamiliar protein pairs or out of distribution samples & State-of-the-art performance on both cross-species and Bernett benchmarks, Uncertainty estimates enable effective filtering of low-confidence predictions & \href{https://github.com/Wang-lab-UCSD/TUnA}{TUnA}\\\cline{2-8}
       & Liu et al. \cite{Liu2023} & Peptide sequences & Binary classification & Bcipep, IEDB & Predicting Linear B-Cell Epitopes & Outperformed deep learning (CNN, BiLSTM) and classical ML models (SVM, Random Forest) across cross-validation and achieved state‑of‑the‑art results when compared to EpitopeVec, iBCE‑EL, LBtope, etc. & \href{https://github.com/liuyf-a/LBCE-XGB}{LBCE-XGB}\\\cline{2-8}
     & Wu et al.~\cite{Wu2024} & Protein sequence, Binding scores & \textit{De novo} generation of binder protein sequences & STRING v11 & Conditions generation by appending a binding score token to the sequence and integrates binding score directly into latent features & Achieved up to 0.98 similarity scores  relative to known binders & N/A\\\cline{2-8}
     &Chu et al.~\cite{ChuRu2024} & Protein sequences and structures & Predicts docked complex structure & DIPS, DB5.5 & Protein-protein docking for both rigid and flexible scenarios & 41–43\% top‑1 success, outperforming other deep-learning models & \href{https://github.com/Graylab/GeoDock}{GeoDock}\\\cline{2-8}
    & Jahn et al. \cite{Jahn2024} & Protein sequences, Disorder annotations & Per-residue binding scores in disordered regions & MobiDB (Training and Testing), DisProt (via CAID2 (Testing)) & Predicts residue-level binding propensity using only single-sequence embeddings from ProtT5, without MSA or
handcrafted features & Outperforms SOTA methods & \href{https://github.com/jahnl/binding_in_disorder}{IDBindT5}\\\hline

\hline
\end{longtable}
\end{center}

\end{landscape} 

\clearpage
\twocolumn
\section{Challenges and Future Directions}
Although Transformer-based models have shown their effectiveness for biological sequences, there are still some pertinent issues that need to be addressed. In this regard, we have summarised some challenges and future directions.
\begin{itemize}
    \item \textbf{Sequence Length}: Protein sequences are usually very long and Transformers have quadratic complexity in attention thereby limiting input length. This may lead to truncation or segmenting resulting in loss of biological context. 
    Sequence chunking involves dividing long sequences into smaller overlapping or non-overlapping segments, which are then processed independently or recurrently. While simple to implement, chunking may compromise the model’s ability to capture global dependencies. Hierarchical representations, as proposed in recent models such as Hierarchical Multi-Modal PLMs~\cite{liu2025}, aim to first learn local residue-level features and then aggregate them into domain- or sequence-level representations, preserving both local and long-range information. Also, sparse attention mechanisms employed in architectures like Longformer~\cite{beltagy2020}, Performers~\cite{choromanski2022}, FlashAttention~\cite{dao2022} and Linformer~\cite{wang2020linformer} for longer sequences can be really helpful as they reduce the computational cost of attention from quadratic to linear or sub-quadratic. 
    More recently, LC‑PLM~\cite{wang2024} has utilised a Structured State Space (SSM) architecture of BiMamba to handle much longer protein sequences with computational efficiency. This model demonstrates a clear architectural strategy to capture long-range dependencies across entire sequences without truncation. They also present LC‑PLM‑G, which incorporates PPI graph context during pretraining to further boost downstream tasks like structure and function prediction.

    \item \textbf{Model Interpretability}: Transformers are often considered as black boxes, but in biological research interpretability is critical for hypothesis generation and validation.
    While attention maps and saliency scores have been used, they may not always correlate with biologically relevant features. Future directions should include systematic benchmarking of interpretability tools (for example, Integrated Gradients, SHAP, attention rollout) against ground-truth annotated sites (for example, binding pockets, PTM sites). Moreover, training with interpretability supervision (for example, using residue-level annotations) may help guide the attention to relevant biological regions.
    
    \item \textbf{Computational Resources}: Large protein Transformers like ESM-2~\cite{Zlin2023} and ProtT5~\cite{Elnaggar2022} require huge GPU/TPU memory and training time which may be a problem for real-time/edge deployment. Instead of general calls for model compression, targeted distillation pipelines (for example, distilling ESM-2 into a smaller model optimised for contact prediction or GO term annotation) should be explored. Additionally, dynamic early exit mechanisms—where the model stops inference once a confident prediction is reached—can reduce computation at inference time. Few-shot adaptation techniques (for example, adapters, LoRA) can also help reduce the training overhead for new tasks.
    
    \item \textbf{Multi-modal Integration}: While many protein models are sequence-based, structure and function annotations provide complementary views. Recent approaches like Uni-Mol2~\cite{ji2024} and AlphaFold3 pretraining pipelines have shown that combining structure and sequence leads to better representations. Future research should systematically investigate how to best align modalities, for example, via cross-attention or contrastive learning—and which modalities provide the greatest marginal gain for specific downstream tasks (for example, enzyme classification, subcellular localisation). Another promising direction is unified pretraining across multiple biological entities (for example, protein + ligand + pathway graphs) to capture functional semantics beyond isolated sequences.

    \item \textbf{Cross-Species Generalization}: Protein Transformer models are often trained and benchmarked on well-characterized species such as humans or model organisms (for example., E. coli, Mus musculus), which may lead to performance degradation when applied to less-studied or phylogenetically distant species. This domain shift limits their utility in biodiversity studies, metagenomics, and non-model species research. One promising direction is domain adaptation—models can be trained with regularization techniques that encourage learning of species-invariant features or finetuned on representative sequences from underrepresented clades. Another approach is zero-shot generalisation, where protein representations are pretrained on taxonomically diverse datasets and evaluated directly on novel species. Additionally, techniques like meta-learning and phylogeny-aware sampling during pretraining can help models better transfer across species by incorporating evolutionary context into the learned representations. Cross-species benchmarks such as those found in Pfam or UniRef50 clusters may further guide the development of models with broader generalisation capability.

    \item \textbf{Robustness and Uncertainty Estimation}: While Transformer models achieve high predictive performance on many protein tasks, they often suffer from poor calibration and overconfidence, especially when encountering out-of-distribution inputs or ambiguous cases such as low-homology sequences or disordered regions. This poses a risk in critical applications like drug discovery or clinical variant interpretation where actionable decisions are based on predictions. To address this, future models should incorporate uncertainty quantification methods, such as Monte Carlo dropout, deep ensembles, or variational Bayesian techniques, which can provide confidence intervals or predictive entropy. Moreover, robust training paradigms, including adversarial training, noise injection, and distributionally robust optimisation can improve generalisation under perturbations. Finally, integrating uncertainty estimates into downstream pipelines (for example, structure prediction, active site detection) can enable more cautious and interpretable decision-making, making protein Transformers more reliable and trustworthy in real-world settings.
    
    \item \textbf{Data quality}: 

Data may affect the validity of the application of Transformers in many aspects: data annotations may be outdated, incomplete, or based on automated methods that propagate errors. Under- or over-representation of classes may create biases in the trained models, and fail to reproduce correctly underrepresented classes. While such issues are shared by most machine learning methods, the field of protein sequences is particularly exposed due to the complexity connected with the annotation of large data sets \cite{kress2023}. While a comprehensive analysis of data quality in this context would be out of scope, we can point out that the literature agrees in recognizing a high annotation quality level to data sets UniProtKB/Swiss-Prot , Pfam and InterPro \cite{Blum2024}.
At the same time, the correct exploitation of statistical and machine learning methods may help in controlling the quality of the results that may depend from the quality of the data: cross-validation, re-sampling, feature selection, use of explanatory techniques, that we have highlighted, when relevant, in the discussion above.

\end{itemize}

\section{Conclusion}
Transformer-based models have brought about a revolution in the field of natural language processing and, then, in bioinformatics; their contributions to processing protein sequences are unprecedented.  In this regard, ProtGPT2 for generating new protein sequences, DistilProtBert for distinguishing between real proteins and their randomly shuffled counterparts etc. are some of the examples to showcase the utility of Transformer-based models in bioinformatics. 
This surge in the use of Transformer-based models has also led to the development of AlphaFold2 and RoseTTAFold for the prediction of secondary and tertiary protein structure.
Applications like protein design, antibody modelling and paratope prediction, enzyme function prediction, as well as MHC binding and immunogenicity prediction are some of the emerging areas where Transformer-based models along with other deep learning techniques can be successfully applied. 
We believe that this review will help the research community working in bioinformatics with new research direction and will help them to come up with new and improved models pertaining to protein design and modelling.









\end{document}